\documentclass[10pt,journal,compsoc]{IEEEtran}
\usepackage[pdftex]{graphicx}
\usepackage[numbers]{natbib}
\usepackage{amssymb}
\usepackage{amsmath}
\usepackage{color}
\usepackage{enumitem}
\usepackage{multirow}
\usepackage{multicol}
\usepackage{graphicx}
\usepackage{lscape}
\usepackage{bbding}
\usepackage{pifont}
\usepackage{booktabs}
\usepackage{diagbox}

\usepackage[switch]{lineno} 
\ifCLASSINFOpdf
   \usepackage[pdftex]{graphicx}
\else
\fi
\begin{document}
%
% paper title
% Titles are generally capitalized except for words such as a, an, and, as,
% at, but, by, for, in, nor, of, on, or, the, to and up, which are usually
% not capitalized unless they are the first or last word of the title.
% Linebreaks \\ can be used within to get better formatting as desired.
% Do not put math or special symbols in the title.
\title{A Survey on Text-to-SQL Parsing: Concepts, Methods, and Future Directions}

\author{Authors
\IEEEcompsocitemizethanks{\IEEEcompsocthanksitem M. Shell was with the Department
of Electrical and Computer Engineering, Georgia Institute of Technology, Atlanta,
GA, 30332.\protect\\
% note need leading \protect in front of \\ to get a newline within \thanks as
% \\ is fragile and will error, could use \hfil\break instead.
E-mail: see http://www.michaelshell.org/contact.html
\IEEEcompsocthanksitem J. Doe and J. Doe are with Anonymous University.}% <-this % stops an unwanted space
\thanks{Manuscript received April 19, 2005; revised August 26, 2015.}}

% note the % following the last \IEEEmembership and also \thanks - 
% these prevent an unwanted space from occurring between the last author name
% and the end of the author line. i.e., if you had this:
% 
% \author{....lastname \thanks{...} \thanks{...} }
%                     ^------------^------------^----Do not want these spaces!
%
% a space would be appended to the last name and could cause every name on that
% line to be shifted left slightly. This is one of those "LaTeX things". For
% instance, "\textbf{A} \textbf{B}" will typeset as "A B" not "AB". To get
% "AB" then you have to do: "\textbf{A}\textbf{B}"
% \thanks is no different in this regard, so shield the last } of each \thanks
% that ends a line with a % and do not let a space in before the next \thanks.
% Spaces after \IEEEmembership other than the last one are OK (and needed) as
% you are supposed to have spaces between the names. For what it is worth,
% this is a minor point as most people would not even notice if the said evil
% space somehow managed to creep in.

% The paper headers
\markboth{IEEE Transactions on Knowledge and Data Engineering}%
{Shell \MakeLowercase{\textit{et al.}}: Bare Demo of IEEEtran.cls for Computer Society Journals}
% The only time the second header will appear is for the odd numbered pages
% after the title page when using the twoside option.
% 
% *** Note that you probably will NOT want to include the author's ***
% *** name in the headers of peer review papers.                   ***
% You can use \ifCLASSOPTIONpeerreview for conditional compilation here if
% you desire.

% The publisher's ID mark at the bottom of the page is less important with
% Computer Society journal papers as those publications place the marks
% outside of the main text columns and, therefore, unlike regular IEEE
% journals, the available text space is not reduced by their presence.
% If you want to put a publisher's ID mark on the page you can do it like
% this:
%\IEEEpubid{0000--0000/00\$00.00~\copyright~2015 IEEE}
% or like this to get the Computer Society new two part style.
%\IEEEpubid{\makebox[\columnwidth]{\hfill 0000--0000/00/\$00.00~\copyright~2015 IEEE}%
%\hspace{\columnsep}\makebox[\columnwidth]{Published by the IEEE Computer Society\hfill}}
% Remember, if you use this you must call \IEEEpubidadjcol in the second
% column for its text to clear the IEEEpubid mark (Computer Society jorunal
% papers don't need this extra clearance.)

% use for special paper notices
%\IEEEspecialpapernotice{(Invited Paper)}

% for Computer Society papers, we must declare the abstract and index terms
% PRIOR to the title within the \IEEEtitleabstractindextext IEEEtran
% command as these need to go into the title area created by \maketitle.
% As a general rule, do not put math, special symbols or citations
% in the abstract or keywords.
\IEEEtitleabstractindextext{%
\begin{abstract}
Text-to-SQL parsing is an essential and challenging task. The goal of text-to-SQL parsing is to convert a natural language (NL) question to its corresponding structured query language (SQL) based on the evidences provided by relational databases. Early text-to-SQL parsing systems from the database community achieved a noticeable progress with the cost of heavy human engineering and user interactions with the systems. In recent years, deep neural networks have significantly advanced this task by neural generation models, which automatically learn a mapping function from an input NL question to an output SQL query. Subsequently, the large pre-trained language models have taken the state-of-the-art of the text-to-SQL parsing task to a new level. In this survey, we present a comprehensive review on deep learning approaches for text-to-SQL parsing. First, we introduce the text-to-SQL parsing corpora which can be categorized as single-turn and multi-turn. Second, we provide a systematical overview of pre-trained language models and existing methods for text-to-SQL parsing. Third, we present readers with the challenges faced by text-to-SQL parsing and explore some potential future directions in this field. 
\end{abstract}

% Note that keywords are not normally used for peerreview papers.
}

\author{Bowen Qin,
        Binyuan Hui,
        Lihan Wang,
        Min Yang,
        Jinyang Li,
        Binhua Li,
        Ruiying Geng,
        Rongyu Cao,
        Jian Sun,
        Luo Si,
        Fei Huang,
        Yongbin Li
\IEEEcompsocitemizethanks{\IEEEcompsocthanksitem B. Qin, L. Wang and M. Yang are with Shenzhen Institutes of Advanced Technology, Chinese Academy of Sciences, Shenzhen, China, 518055. B. Qin and L. Wang are also with University of Chinese Academy of Sciences, Beijing, China, 101408.  \protect\\
% note need leading \protect in front of \\ to get a newline within \thanks as
% \\ is fragile and will error, could use \hfil\break instead.
E-mail: \{lh.wang1, bw.qin, min.yang\}@siat.ac.cn
\IEEEcompsocthanksitem B. Hui, B. Li, R. Geng, R. Gao, J. Sun, L. Si, F. Huang and Y. Li are with Alibaba Group, Beijing, China. \protect\\
E-mail: \{binyuan.hby, binhua.lbh, ruiying.gry, caorongyu.cry, jian.sun, luo.si, f.huang, shuide.lyb\}@alibaba-inc.com
\IEEEcompsocthanksitem J. Li is with The University of Hong Kong, Hong Kong. \protect\\
E-mail: jl0725@connect.hku.hk
}% <-this % stops an unwanted space
\thanks{Min Yang and Yongbin Li are corresponding authors.}
}

% make the title area
\maketitle

% To allow for easy dual compilation without having to reenter the
% abstract/keywords data, the \IEEEtitleabstractindextext text will
% not be used in maketitle, but will appear (i.e., to be "transported")
% here as \IEEEdisplaynontitleabstractindextext when the compsoc 
% or transmag modes are not selected <OR> if conference mode is selected 
% - because all conference papers position the abstract like regular
% papers do.
\IEEEdisplaynontitleabstractindextext
% \IEEEdisplaynontitleabstractindextext has no effect when using
% compsoc or transmag under a non-conference mode.

% For peer review papers, you can put extra information on the cover
% page as needed:Neural Approaches for Natural Language Interfaces to Databases:

% \ifCLASSOPTIONpeerreview
% \begin{center} \bfseries EDICS Category: 3-BBND \end{center}
% \fi
%
% For peerreview papers, this IEEEtran command inserts a page break and
% creates the second title. It will be ignored for other modes.
\IEEEpeerreviewmaketitle

\IEEEraisesectionheading{\section{Introduction}\label{sec:introduction}}
With the popularity of electronic devices, tables have become the mainstream to store large structural data from various resources (e.g., webpages, databases and spreadsheets), which represent the data as a grid-like format of rows and columns so that users can easily inquire the patterns and discover insights from data. Although the tables can be efficiently accessed by skilled professionals via the handcrafted structured query languages (SQLs), a natural language (NL) interface can facilitate the ubiquitous relational data to be accessed by a wider range of non-technical users \cite{wang2022proton}. 
Therefore, text-to-SQL parsing, which aims to translate NL  questions to machine-executable SQLs, has attracted noticeable attention from both  industrial and academic communities. It can empower non-expert users to effortlessly query tables and plays a central role in various real-life applications such as intelligent customer service, question answering, and robotic navigation.

Early text-to-SQL parsing work \cite{zelle1996learning} from the Database community made a noticeable progress with the cost of heavy human engineering and user interactions with the systems. It is difficult, if not impossible, to design SQL templates in advance for various scenarios or domains. 
In recent years, recent advances of deep learning and the availability of large-scale training data have significantly improve text-to-SQL parsing by neural generation models. A typical neural generation method is the sequence-to-sequence (Seq2Seq)~\cite{sutskever2014sequence} model,  which automatically learns a mapping function from the input NL question to the output SQL under encoder-decoder schemes. 
The key idea is to construct an encoder to understand the input NL questions together with related table schema and leverage a grammar-based neural decoder to predict the target SQL. 
The Seq2Seq based approaches have become the mainstream for text-to-SQL parsing mainly because they can be trained in an end-to-end way and reduce the need for specialized domain knowledge. 

So far, various neural generation models have been developed to improve the encoder and the decoder respectively. On the encoder side, several general neural networks are widely used to globally reason over natural language query and database schema. IRNet \cite{guo2019towards} encoded the question and the table schema separately with bi-directional LSTM \cite{hochreiter1997long} and self-attention mechanism \cite{vaswani2017attention}. RYANSQL \cite{choi2021ryansql} employed convolutional neural network \cite{o2015introduction} with dense connection \cite{yoon2018dynamic} for question/schema encoding.
With the advance of pre-trained language models (PLMs), SQLova \cite{hwang2019comprehensive} first proposed to leverage the pre-trained language models (PLMs) such as BERT \cite{devlin2018bert} as the base encoder. 
RATSQL \cite{wang2019rat}, SADGA \cite{cai2021sadga} and LGESQL \cite{cao2021lgesql} adopted graph neural network to encode the relational structure between the database schema and a given question.
On the decoder side, there are two categories of SQL generation approaches including the sketch-based methods and the generation-based methods. Specifically, the sketch-based methods \cite{xu2017sqlnet, hwang2019comprehensive, hui2021improving} decompose the SQL generation procedure into sub-modules, where each sub-module corresponds to the type of the prediction slot to be filled. These sub-modules are later gathered together to generate the final SQL query.
To enhance the performance of the generated SQL logic form, the generation-based methods \cite{guo2019towards,wang2019rat,cao2021lgesql,huang2021relation} usually decoded the SQL query as an abstract syntax tree in the depth-first traversal order by employing an LSTM \cite{hochreiter1997long} decoder.

\begin{figure*}[!t]
    \centering
    \includegraphics[width=18cm]{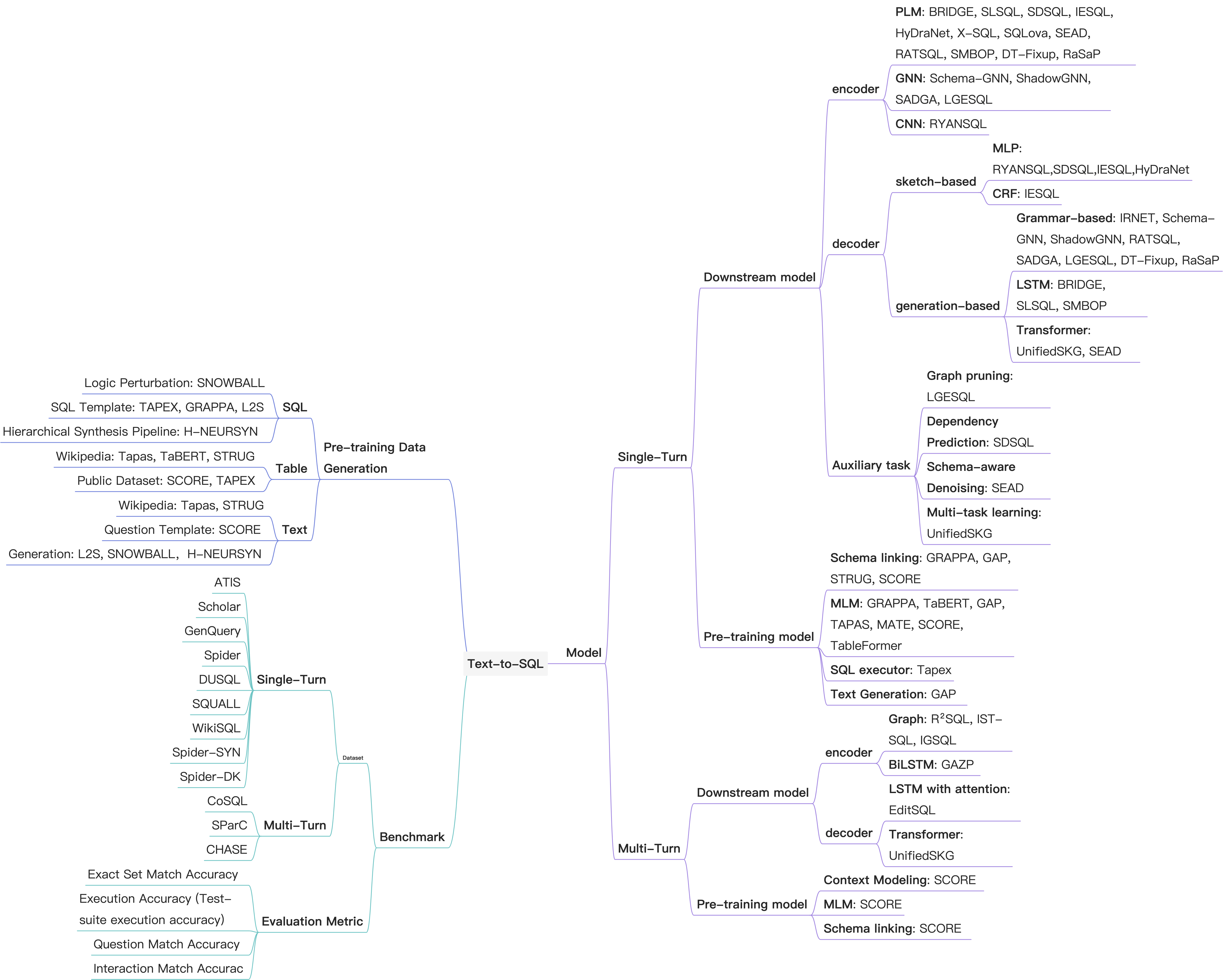}
    \caption{The comprehensive overview of the text-to-SQL parsing datasets, the pre-training tabular language models, the downstream text-to-SQL parsing approaches.}
    \label{fig:my_label}
\end{figure*}

In parallel, PLMs have proved to be powerful in enhancing text-to-SQL parsing and yield state-of-the-art performances, which benefit from the rich linguistic knowledge in large-scale corpora. However, as revealed in previous works, there are intrinsic differences between the distribution of tables and plain texts. Directly fine-tuning the PLMs trained on large-scale plain texts to downstream text-to-SQL parsing hinders the models from effectively modeling the relational relational structure in question/schema, and thus leads to sub-optimal performances. Current studies to alleviate the above limitation attempt to build Tabular Language Models (TaLMs) by directly encoding tables and texts, which show improved results on downstream text-to-SQL parsing tasks. For example, TaBERT \cite{yin2020tabert} jointly encoded texts and tables with masked language modeling (MLM) and masked column prediction (MCP) respectively, which was trained on a large corpus of tables and their corresponding English contexts.
TaPas \cite{herzig-etal-2020-tapas} extended BERT \cite{devlin2018bert} by using additional positional embeddings to encode tables. In addition,  two classification layers are applied to choose table cells and aggregation operators which operate on the table cells. 
Grappa \cite{yu2020grappa} introduced a grammar-augmented pre-training framework for table semantic parsing, which explores the schema linking in table semantic parsing by encouraging the model to capture table schema items which can be grounded to logical form constituents. Grappa  achieved the state-of-the-art performances for text-to-SQL parsing. 

\vspace{0.2cm}

\noindent \textbf{Contributions of this survey.} This manuscript aims at providing a comprehensive review of the literature on text-to-SQL parsing as shown in Fig. \ref{fig:my_label}. By providing this survey, we hope to provide a useful resource for both academic and industrial communities. 
First, we introduce the experimental datasets and present a taxonomy that classifies the representative text-to-SQL approaches.
Moreover, we present readers with the challenges faced by text-to-SQL parsing and explore some potential future directions in this field.

This manuscript is organized as follows. In Section 2, we define the text-to-SQL parsing formally and introduce the official evaluation metrics. Section 3 presents the main scenarios (single-turn and multi-turn utterances) and the corresponding datasets for text-to-SQL parsing. We introduce the representative pre-training, encoding and decoding techniques for text-to-SQL parsing in Section 4 and Section 5 respectively. Section 6 concludes this manuscript and outlines the future directions, followed by the references.

\begin{table*}[!t]
\caption{The notations used in this manuscript.}
    \centering
    \begin{tabular}{c|l}
    \toprule
      Symbol  & Description \\
    \midrule
        $\mathcal{S}$   & Sequence of database schema tokens, which consists of tables and columns. \\ 
        $\mathcal{T}$   & Sequence of table tokens.\\ 
        $ \mathcal{C}$  & Sequence of column tokens.\\
        $ \mathcal{Q}$  & Sequence of question tokens.\\
        $ q $           & Question token. \\
        $ t $           & Table token. \\
        $ c $           & Column token. \\
        $ | | $         & Length of tokens. \\
        $ X $           & Input of text-to-SQL model, which consists of question and schema. \\
        $ Y $           & Output of text-to-SQL model, referring to SQL query.\\
        $ I $           & Input sequence of encoder, which consists of special token, question token and schema tokens.\\
        $\texttt{[CLS]}, \texttt{[SEP]} $ & Special token of PLMs. \\
        $ \mathbf{u} $  & Graph node embedding vector. \\
        $ \mathbf{r} $  & Relation embedding vector. \\
        $ W_{K} $       & Weight matrix of key vectors, which are used to calculate the attention score. \\
        $ W_{Q} $       & Weight matrix of query vectors, which are used to calculate the attention score. \\
        \bottomrule
    \end{tabular}
    \label{tab:notation}
\end{table*}

\section{Background}
In this section, we first provide a formal problem definition of text-to-SQL parsing. Then, we describe the official evaluation metrics for verifying the text-to-SQL parsers. Finally, we introduce the benchmark corpora used for training the neural text-to-SQL parsers.

\subsection{Task Formulation}
\label{sec:task-definition}
Text-to-SQL (T2S) parsing aims to convert a natural language (NL) question under database items  to its corresponding structured query language (SQL) that can be executed against a relational database. 
As shown in Table \ref{tab:notation}, we provide formal notation to normalise task definitions.
Generally, existing T2S parsing approaches can be categorized into single-turn (context-independent) and multi-turn (context-dependent) settings. 
Formally, for the single-turn T2S parsing setting, given a NL question $Q$ and the corresponding database schema $\mathcal{S}=\langle\mathcal{T}, \mathcal{C}\rangle$, our goal is to generate a SQL query $Y$. To be specific, the question $Q = \left\{q_{1}, q_{2}, \cdots, q_{|Q|} \right\}$ is a sequence of $|Q|$ tokens. The database schema consists of $|\mathcal{T}|$ tables $\mathcal{T}=\left\{t_{1}, t_{2}, \cdots, t_{|\mathcal{T}|} \right\}$  and $|\mathcal{C}|$ columns $C=\left\{c_{1}, c_{2}, \cdots, c_{|\mathcal{C}|} \right\}$. Each table $t_i$ is described by its name that contains multiple words $[t_{i,1}, t_{i,2}, \cdots, t_{i,|t_i|}]$. Each column $c_j^{t_i}$ in table $t_i$ is represented by words (a phrase) $[c_{j,1}^{t_{i}}, c_{j,2}^{t_{i}}, \cdots, c_{j,|c_j^{t_i}|}^{t_{i}}]$. We denote the whole input as $X = \langle\mathcal{Q}, \mathcal{S}\rangle$. 

For the multi-turn T2S parsing setting, we aims to convert a sequence of NL questions to the corresponding SQL queries, where the NL questions may contain ellipsis and anaphora that refers to earlier items in the previous NL questions. 
Formally, let $U=\{U_1,\ldots, U_T\}$ denote a sequence of utterances  with $T$ turns, where $U_t=(X_t,Y_t)$ represents the $t$-th utterance which is the combination of a NL question $X_i$ and a SQL query $Y_i$. In addition, there is corresponding database schema $\mathcal{S}$.  At the $t$-th turn, the goal of multi-turn T2S parsing is to produce the SQL query $Y_t$ conditioned on the current NL question $X_t$, the historical utterances $\{U_i\}_{i=1}^{t-1}$, and the database schema $\mathcal{S}$. 

\begin{table*}[!t]
\centering
\caption{The statistics of the T2S datasets. ``\#'' denotes the number of the corresponding units.}
    \begin{tabular}{c|ccccc|ccccc}
        \toprule
        \textbf{Dataset}                                    &Single-Turn        &Multi-Turn         &Cross-domain      &Robustness  &Languages  &\#Question     &\#SQL      &\#DB       &\#Domain   &\#Table  \\ 
        \midrule
        GenQuery \cite{zelle1996learning}                   &\ding{51}          &                   &                  &            &en  &880            &247        &1          &1          &6     \\
        Scholar \cite{iyer2017learning}                     &\ding{51}          &                   &                  &            &en  &817            &193        &1          &1          &7     \\
        WikiSQL \cite{zhong2017seq2sql}                     &\ding{51}          &                   &                  &            &en  &80654          &77840      &26521      &-          &1     \\
        Spider \cite{yu2018spider}                          &\ding{51}          &                   &\ding{51}         &            &en  &10181          &5693       &200        &138        &1020     \\
        Spider-SYN  \cite{spiderSYN}                        &\ding{51}          &                   &\ding{51}         & \ding{51}  &en  &7990           &4525       &166        &-          &876      \\
        Spider-DK \cite{spiderDK}                           &\ding{51}          &                   &\ding{51}         & \ding{51}  &en  &535            &283        &10         &-          &48       \\
        Spider-SSP  \cite{shaw2020compositional}            &\ding{51}          &                   &                  &            &en  &-           &-      &-      &-          &-    \\
        CSpider \cite{min2019pilot}                         &\ding{51}          &                   &\ding{51}         &            &zh  &10181          &5693       &200        &138        &1020     \\
        SQUALL \cite{min2019pilot}                          &\ding{51}          &                   &\ding{51}         &            &en  &15620          &11276      &2108       &-          &2108     \\
        DuSQL \cite{wang2020dusql}                          &\ding{51}          &                   &\ding{51}         &            &zh  &23797          &23797      &200        &-          &820      \\ 
        ATIS \cite{price1990evaluation,dahl1994expanding}   &                   &\ding{51}          &                  &            &en  &5418           &947        &1          &1          &27    \\
        SparC \cite{yu2019sparc}                            &                   &\ding{51}          &\ding{51}         &            &en  &4298           &12726      &200        &138        &1020     \\
        CoSQL \cite{yu2019cosql}                            &                   &\ding{51}          &\ding{51}         &            &en  &3007           &15598      &200        &138        &1020     \\
        CHASE \cite{guo2021chase}                           &                   &\ding{51}          &\ding{51}         &            &zh  &5489           &17940      &280        &-          &1280     \\
        
        \bottomrule
    \end{tabular}
    \label{tab:datasets}
\end{table*}

\subsection{Evaluation metrics}
The text-to-SQL (T2S) parsers are generally evaluated by comparing the generated SQL queries against the ground-truth SQL answers. 
Concretely, there are two types of evaluation metrics that are used for evaluating the single-turn T2S setting, including exact set match accuracy (EM) and execution accuracy (EX) \cite{yu2018spider}. For the multi-turn T2S setting, question match accuracy (QM) and interaction match accuracy (IM) \cite{yu2019sparc} are commonly employed.

\subsubsection{Single-turn T2S Evaluation}
\vspace{0.2cm}
\noindent \textbf{Exact Set Match Accuracy (EM)}
The exact set match accuracy (without values) is calculated by comparing the ground-truth SQL query and the predicted SQL query. Both ground-truth and predicted queries are parsed into normalized data structures which have the following SQL clauses such as
\texttt{SELECT} • \texttt{GROUP BY} •  \texttt{WHERE} • \texttt{ORDER BY} • \texttt{KEYWORDS} (including all SQL keywords without column names and operators).

We treat the predicted SQL query as correct only if all of the SQL clauses are correct by a set comparison as follows:
\begin{equation}
    score(\hat{Y}, Y) = \left\{\begin{aligned}
        1, \quad \hat{Y} = Y \\ 0, \quad \hat{Y} \neq {Y}
    \end{aligned}
    \right.
\end{equation}
where $\hat{Y} = \{ (\hat{k}^{i}, \hat{v}^{i}), i \in (1, m)\}$ and $Y = \{ (k^{i}, v^{i}), i \in (1, m)\}$ denote the component sets of the predicted SQL query and the ground-truth query respectively. Here, $k$ stands for a SQL clause and $v$ is the corresponding value of the clause.  $m$ is the number of components.
Formally, the exact set match accuracy is calculated by:
\begin{equation}
    {\rm EM} = \frac{ \Sigma_{n=1}^{N} score(\hat{Y}_n, Y_n) } { N },
\end{equation}
where $N$ denotes the total number of samples.
EM evaluates the model performance by strictly comparing differences in SQL, but human's SQL annotations are often biased since a NL question may correspond to multiple SQL queries. 

\vspace{0.2cm}
\noindent \textbf{Execution Accuracy (EX)}
Execution accuracy (with values) is calculated by comparing the output results of executing the ground-truth SQL query and the predicted SQL query on the database contents shipped with the test set.
We treat the predicted query as correct only if the results of executing the predicted SQL query $\hat{V}$ and the ground-truth SQL query $V$ are same:
\begin{equation}
    score(\hat{V}, V) = \left\{\begin{aligned}
        1, \quad \hat{V} = V \\ 0, \quad \hat{V} \neq {V}
    \end{aligned}
    \right.
\end{equation}
Similarly with EM, the EX is calculated by:
\begin{equation}
    {\rm EX} = \frac{ \Sigma_{n=1}^{N} score(\hat{Y}_{n}, Y_{n})} { N },
\end{equation}
To avoid false positives and false negatives caused by SQL execution on finite size databases, the test-suite execution accuracy \cite{ruiqi20} extends the execution to multiple database instances per schema. 
Concretely, the test-suite distills a small database from random generated databases to achieve high code coverage. In this way, we can provide the best approximation of semantic accuracy.
%The contents of these instances are optimized to lower the number of false positives and to provide the best approximation of semantic accuracy.

\subsubsection{Multi-turn T2S Evaluation}
Given a multi-turn setting, there are a total of $P$ question sequences, where each sequence contains $O$ rounds and a total of $M = P \times O$ questions.

\vspace{0.2cm}
\noindent \textbf{Question Match Accuracy (QM)}
The question match accuracy is calculated as the EM score over all questions.
Its value is $1$ for each question only if all predicted SQL clauses are correct. We first calculate the EM score for each question as follows: 
\begin{equation}
    score(\hat{Y}, Y) = \left\{\begin{aligned}
        1, \quad \hat{Y} = Y \\ 0, \quad \hat{Y} \neq {Y}
    \end{aligned}
    \right.
\end{equation}
where $\hat{Y}$ and $Y$ are the predicted and ground-truth SQL queries, respectively. Then, the question match accuracy is calculated by:
\begin{equation}
    {\rm QM} = \frac{ \Sigma_{m=1}^{M} score(\hat{Y}_{m}, Y_{m}) } { M },
\end{equation}
where $M$ denotes the total number of questions. 

\vspace{0.2cm}
\noindent \textbf{Interaction Match Accuracy (IM)}
The interaction match accuracy is calculated as the EM score over all interactions (question sequences). The score of each interaction is $1$  only if all the questions within the interaction are correct. Formally, the score for each interaction is defined as follows:
\begin{equation}
    interaction = \left\{\begin{aligned}
        1, \quad \prod_{i=1}^{o} score(\hat{Y}_i,Y_i) = 1 \\ 0, \quad \prod_{i=1}^{o} score(\hat{Y}_i,Y_i) = 0
    \end{aligned}
    \right.
\end{equation}
where $o$ is the number of turns in each interaction.
Then, the IM score is calculated by:
\begin{equation}
    {\rm IM} = \frac{ \Sigma_{p=1}^{P} {interaction}_{p} } { P }.
\end{equation}
where $P$ is the total number of interactions.

\subsection{Datasets}
High-quality corpora are essential for learning and evaluating the text-to-SQL (T2S) parsing systems. In the following, we summarize extensively-used datasets into two primary categories: the single-turn T2S corpora with single-turn (stand-alone) questions and the multi-turn T2S corpora with multi-turn sequential questions. 

\subsubsection{Single-Turn T2S Corpora}

\vspace{0.3cm}
\noindent \textbf{GenQuery~~}
The GenQuery \cite{zelle1996learning} dataset is a collection 880 NL questions for querying a database of US geographical facts (denoted as Geobase). A relational database schema and SQL queries are constructed over Geobase for 700 questions. Afterwards, the remaining NL questions are further annotated by \cite{iyer2017learning}, following the widely used 600/280 training/test split \cite{zettlemoyer2012learning}.

\vspace{0.3cm}
\noindent \textbf{Scholar~~}
The Scholar \cite{iyer2017learning} dataset is a collection of 816 NL questions annotated with SQL queries, where 600 NL questions are used for training and 216 questions are used for testing. An academic database consisting of academic papers is provided to executed the SQL queries. 

\vspace{0.3cm}
\noindent \textbf{WikiSQL~~}
The original WikiSQL \cite{zhong2017seq2sql} dataset is a collection of 80,654 hand-crafted NL question and SQL query pairs along with the corresponding SQL tables extracted from 24,241 HTML tables on Wikipedia. In particular, for each selected table six SQL queries are generated following the SQL templates and rules. Then, for each SQL query, a crude NL question is annotated using templates via crowd-sourcing on Amazon Mechanical Turk. The WikiSQL dataset contains much more instances and tables than ATIS \cite{price1990evaluation,dahl1994expanding}, GenQuery \cite{zelle1996learning} and Scholar \cite{iyer2017learning}. In addition, the WikiSQL dataset is more challenging than previous T2S corpora since WikiSQL spans over a large number of tables and the text-to-SQL parsers should generalize to  not only  new queries but also new table schema. The instances in WikiSQL can be randomly split into training/validation/testing sets, such that each table is involved in exactly
one set.

\vspace{0.3cm}
\noindent \textbf{Spider~~}
The Spider \cite{yu2018spider} dataset  is a large-scale benchmark for cross-domain text-to-SQL parsing. Spider contains 10,181 NL questions and 5,693 unique SQL queries over 200 databases belonging to 138 different domains. Different from the prior T2S datasets that contain tables from the same domain, the Spider dataset contains complex NL questions and SQL queries spanning over multiple databases and domains. 
In addition, the SQL queries in the Spider dataset can be further divided into four levels based on the difficulty of the SQL queries: easy, medium, hard, extra hard, which can be used to better evaluate the model performance on different queries. 
Finally, the Spider dataset is randomly split into 7,000 instances for training, 1,034 instances for development, and 2,147 instances for testing.

\vspace{0.3cm}
\noindent \textbf{Spider-Syn~~} 
The Spider-Syn \cite{spiderSYN} dataset is another challenging variant of Spider \cite{yu2018spider}, which modifies the NL questions from Spider \cite{yu2018spider} by replacing the schema-related words with the corresponding synonyms. In this way, the explicit alignments between the words in NL questions and the tokens in table schemas are eliminated, which make the schema linking more challenging for text-to-SQL parsing. Spider-Syn \cite{spiderSYN} is composed of 7000 training instances and 1034 development instances. Note that Spider-Syn  \cite{spiderSYN} has no test set since the original Spider \cite{yu2018spider} does not release the test set publicly.

\vspace{0.3cm}
\noindent \textbf{Spider-DK~~} 
The Spider-DK \cite{spiderDK} dataset is a challenging variant of the Spider \cite{yu2018spider} development set, which can be used to better investigate the generalization ability of existing text-to-SQL models in understanding the domain knowledge. Spider-DK is constructed by adding domain knowledge that reflects real-world question paraphrases to some NL questions from the Spider development set. Concretely, it consists of 535 NL-SQL pairs, where $265$ NL-SQL pairs are modified by adding  domain knowledge while the rest $270$ NL-SQL pairs remain the same as in the original Spider dataset. It is noteworthy that Spider-DK is smaller than the original Spider development set since not every instance can be easily modified to add domain knowledge.

\vspace{0.3cm}
\noindent \textbf{Spider-SSP~~}
The Spider-SSP \cite{shaw2020compositional} refers to the compositional generalization version of the Spider \cite{yu2018spider} dataset. This ability to generalize to novel combinations of the elements observed during training is referred to as compositional generalization. A new train and test split of the Spider dataset is proposed based on Target Maximum Compound Divergence (TMCD) \cite{shaw2020compositional}. Spider-SSP consists of $3,282$ training instances and $1,094$ testing instances, and the databases are shared between  the training and testing instances.

\vspace{0.3cm}
\noindent \textbf{CSpider~~} 
The CSpider \cite{min2019pilot} dataset is a Chinese variant of Spider \cite{yu2018spider} by translating the English NL questions in Spider into Chinese. Similar to Spider, CSpider consists of the same question-SQL pairs as in the Spider dataset.

\vspace{0.3cm}
\noindent \textbf{SQUALL~~}
The SQUALL \cite{shi2020potential} dataset is an extension of {WIKITABLEQUESTIONS} \cite{pasupat2015compositional}, which enriches the $11,276$ samples from the training set of {WIKITABLEQUESTIONS} by providing hand-crafted annotations including both SQL queries and the labeled alignments between NL question tokens as well as the corresponding SQL fragments. In total, SQUALL contain $15,620$ instances, which are split into $9,030$ instances for training, $2,246$ instances for validation, and $4,344$ instances for testing. 

\vspace{0.3cm}
\noindent \textbf{DuSQL~~}
The DuSQL \cite{wang2020dusql} dataset is a larges-scale Chinese corpus for cross-domain text-to-SQL parsing, which consists of $23,797$ NL-SQL pairs along with $200$ databases and $813$ tables belonging to more than 160 domains. Different from most previous corpora that are manually annotated, the SQL queries in DuSQL are automatically generated via production rules from the grammar.

\subsubsection{Multi-Turn T2S Corpora}
\noindent \textbf{ATIS~~}
The original ATIS \cite{price1990evaluation,dahl1994expanding} dataset is a collection of user questions asking for flight information on airline travel inquiry systems along with a relational database that contains information about cities, airports, flights, and so on. Most of the posted questions can be answered by querying the database with SQL queries. Since the original SQL queries are inefficient to be executed by using the IN clauses, the SQL queries are further modified by \cite{iyer2017learning} while keeping the output of the SQL queries unchanged. In total, there are 5,418 NL utterances with corresponding executable SQL queries, where 4,473 utterances for training, 497 for development and 448 for testing. 

\vspace{0.3cm}
\noindent \textbf{SParC~~}
The SParC \cite{yu2019sparc} dataset is a large-scale cross-domain context-dependent text-to-SQL corpus, which contains about 4.3k question sequences including 12k+ question-SQL pairs along with 200 complex databases belonging to $138$ domains. SParC is built on the Spider \cite{yu2018spider}, where each question sequence is based on a question from Spider by asking inter-related questions. After obtaining the sequential questions, a SQL query is manually annotated for each question. Following Spider, SParC is split into training, development and test sets with a ratio of 7:1:2, such that each database appears in only one set. 
    
\vspace{0.3cm}
\noindent \textbf{CoSQL~~}
The CoSQL \cite{yu2019cosql} dataset is the first large-scale
cross-domain conversational text-to-SQL dataset created under the WOZ setting, which consists of about 3k dialogues including 30k+ turns and 10k+ corresponding SQL queries along with $200$ complex databases belonging to $138$ domains. In particular, each conversation simulates a DB query scenario where the annotators working as DB users issue NL questions to retrieve answers with SQL queries. The sequential NL questions can be used to clarify historical ambiguous questions or notify users of unanswerable questions. Similar to Spider \cite{yu2018spider} and SParC \cite{yu2019sparc} , CoSQL is also split into training, development and test sets with a ratio of 7:1:2, such that each database appears in only one set. 

\vspace{0.3cm}
\noindent \textbf{CHASE~~}
The CHASE \cite{guo2021chase} dataset is a large-scale context-dependent chinese text-to-SQL corpus, which is composed of $5,459$ coherent question sequences including $17,940$ questions with their SQL queries. The context-dependent question-SQL pairs span over $280$ relational databases. CHASE has to variants: CHASE-C and CHASE-T. Specifically, CHASE-C collects $120$ Chinese relational databases from DuSQL \cite{wang2020dusql} and creates $2003$ question sequences as well as their SQL queries from scratch. CHASE-T is created by translating the $3456$ English questions sequences and $160$ databases from SParC \cite{yu2019sparc} into Chinese. The CHASE dataset is split into $3,949$/$755$/$755$ samples for training, validation and testing, such that a database appears in solely one set. 
\section{Single-turn T2S Parsing Approaches}

\begin{table*}[!ht]
\centering
\caption{The representative downstream text-to-SQL parsing approaches. EM denotes the exact match accuracy on the Spider \cite{yu2018spider} data for the latest submissions. The columns denote the most important architecture decisions (SL - Schema Linking, DSL - Domain Specific Language).}
\label{tab:downstream-methods}
\resizebox{\textwidth}{!}{
\begin{tabular}{l|cccc|ccccccc|c|c|c|c}
\toprule
\multirow{2}{*}{\textbf{Model}}    &   
\multicolumn{4}{c|}{\textbf{Encoder}}   &   
\multicolumn{7}{c|}{\textbf{Decoder}}   &
\multirow{2}{*}{\textbf{EM Dev}} & 
\multirow{2}{*}{\textbf{EM Test}} &
\multirow{2}{*}{\textbf{EX Dev}} & 
\multirow{2}{*}{\textbf{EX Test}} \\

\multicolumn{1}{c|}{}                   &SL         &LSTM       &Transformer&GNN        &LSTM       &Transformer       &Grammar    &Sketch     &DSL       &Constrained Decoding   & Re-Ranking    &   &   & & \\
\midrule
Seq2Seq baseline \cite{yu2018spider}    &           &\ding{51}  &           &           &\ding{51}  &                   &           &           &           &   &           &1.8    &4.8    & - & -   \\
TypeSQL\cite{yu2018typesql}             &\ding{51}  &\ding{51}  &           &           &           &                   &           &\ding{51}  &           &   &           &8.9    &8.2    &- & -   \\
SyntaxSQLNet \cite{yu2018syntaxsqlnet}  &           &           &           &\ding{51}  &\ding{51}  &                   &\ding{51}  &           &           &   &           &25.0   &-      &- &    \\
GNN \cite{bogin2019representing}        &\ding{51}  &           &           &\ding{51}  &\ding{51}  &                   &\ding{51}  &           &           &   &           &51.3   &-      &- &-    \\
EditSQL\cite{zhang2019editing}          &           &\ding{51}  &\ding{51}  &           &\ding{51}  &                   &           &           &           &   &           &57.6   &53.4   &- &-    \\
Bertrand-DR\cite{kelkar2020bertrand}    &           &\ding{51}  &\ding{51}  &           &\ding{51}  &                   &           &           &           &   &\ding{51}  &58.5   &-      &- &-   \\
IRNet\cite{guo2019towards}              &\ding{51}  &           &           &           &\ding{51}  &                   &\ding{51}  &           & \ding{51} &   &           &61.9   &54.7   &- &-   \\
RYANSQL\cite{choi2021ryansql}           &           &           &           &           &           &                   &           &\ding{51}  &           &   &           &66.6   &58.2   &- &-   \\
BRIDGE\cite{lin2020bridging}            &\ding{51}  &           &\ding{51}  &           &\ding{51}  &                   &\ding{51}  &           &           &   &           &70.0   &65.0   &70.3 & 68.3  \\
RATSQL\cite{wang2019rat}                &\ding{51}  &           &\ding{51}  &           &\ding{51}  &                   &\ding{51}  &           &           &   &           &69.7   &65.6   &- &-   \\
SMBOP  \cite{rubin2020smbop}            &\ding{51}  &           &\ding{51}  &           &           &\ding{51}          &\ding{51}  &           &           &   &           &69.5   &71.1   & 75.0 &71.1   \\
ShadowGNN  \cite{chen2021shadowgnn}     &\ding{51}  &           &\ding{51}  &           &\ding{51}  &                   &\ding{51}  &           &\ding{51}  &   &           &72.3   &66.1   &- &-   \\
RaSaP\cite{huang2021relation}           &\ding{51}  &           &\ding{51}  &           &           &\ding{51}          &\ding{51}  &           &           &   &           &74.7   &69.0   &- &70.0   \\
SADGA\cite{cai2021sadga}                &\ding{51}  &           &           &\ding{51}  &\ding{51}  &                   &\ding{51}  &           &           &   &           &73.1   &70.1   &- & -  \\
DT-Fixup\cite{xu2020optimizing}         &\ding{51}  &           &\ding{51}  &           &           &\ding{51}          &           &           &           &   &           &75.0   &70.9   &- & -  \\
T5-Picard\cite{scholak2021picard}       &           &           &\ding{51}  &\ding{51}  &           &\ding{51}          &           &           &           &\ding{51}&&75.5  &71.9   &79.3 &75.1   \\
LGESQL\cite{cao2021lgesql}              &\ding{51}  &           &           &\ding{51}  &\ding{51}  &                   &\ding{51}  &           &           &   &           &75.1   &72.0  &- & -  \\
S$^2$SQL\cite{hui2022s}                 &\ding{51}  &           &           &\ding{51}  &\ding{51}  &                   &\ding{51}  &           &           &   &           &76.4   &72.1  &- &  -  \\ 
\hline
\end{tabular}
}
\end{table*}

Deep learning has long been dominant in the field of text-to-SQL parsing, yielding state-of-the-art performances. 
In this manuscript, we provide a comprehensive review of recent neural network-based approaches for text-to-SQL parsing.
A typical neural text-to-SQL method is usually based on the sequence-to-sequence (Seq2Seq) model \cite{sutskever2014sequence}, in which an encoder is devised to capture the semantics of the NL question with a real-valued vector and a decoder is proposed to generate the SQL query token by token based on the encoded question representation. 
As illustrated in Table \ref{tab:downstream-methods}, we divide the downstream text-to-SQL parsing methods into several primary categories based on the \textbf{encoder} and the \textbf{decoder}. 
Next, we describe each category of the text-to-SQL parsing methods in detail.  

\subsection{Encoder}
The first goal of the encoder is to learn \textbf{input representation}, jointly representing the NL question and table schema representations. The second goal of the encoder is  to perform \textbf{structure modelling}, since the text-to-SQL parsing task is in principle a highly structured task. %where language is structured, schemas are structured, and the complex interactions between language and schema are structured.

\subsubsection{Input Representation}
As stated in Section \ref{sec:task-definition}, there are two types of input information to be considered for text-to-SQL parsing: the NL question and the table schemas, which are jointly represented by the encoder. 
Generally, the input representation learning methods can be divided into two primary categories, including LSTM-based \cite{hochreiter1997long,schuster1997bidirectional} and Transformer-based \cite{vaswani2017attention} methods.

\vspace{0.3cm}
\noindent \textbf{LSTM-based Methods} \quad
Motivated by the significant success in text representation learning, LSTM-based methods \cite{hochreiter1997long,schuster1997bidirectional} are widely used to learn contextualized representations of input NL question and table schema, which are then passed into the decoder for generating SQL query. 
TypeSQL \cite{yu2018typesql}, Seq2SQL \cite{zhong2017seq2sql} and SyntaxSQLNet \cite{yu2018syntaxsqlnet} work on the stand-alone question-SQL pairs and adopt the bidirectional LSTM (Bi-LSTM) to learn semantic representations of the input sequence which is the concatenation of the NL question and the column names. 
IRNet \cite{guo2019towards} encodes the NL question and the table schema by using two separate Bi-LSTM encoders. 
In particular, the two Bi-LSTM encoders take as input the word embeddings and the corresponding schema linking type embeddings, where the schema linking type embeddings are obtained by applying n-gram string matching to identify the table and column names mentioned in the NL question. 

\vspace{0.3cm}
\noindent \textbf{Transformer-based Methods} \quad
Recently, the Transformer-based \cite{vaswani2017attention} models have shown state-of-the-art performances on text representation learning for multiple natural language processing (NLP) tasks. There are also several text-to-SQL parsing methods such as SQLova~\cite{hwang2019comprehensive} and SLSQL~\cite{lei2020re} that extend BERT~\cite{devlin2018bert} and RoBERTa~\cite{liu2019roberta} for encoding the NL question together with the table and column headers. Generally, the Transformer-based encoders follow a three-step procedure. 

\textbf{First}, the NL question and the database schema are concatenated and taken as the integrated input sequence of the encoder. Formally, the input sequence can be formulated as $\mathcal{I}= ({\rm [CLS]};q_1;\dots;q_{|Q|};{\rm [SEP]}; s_1;{\rm [SEP]};\dots;{\rm [SEP]}; s_{|\mathcal{T}| + |\mathcal{C}|})$, where ${\rm [CLS]}$ and ${\rm [SEP]}$ indicate the pre-defined special tokens as in \cite{devlin2018bert}. The input sequence can be extended to the multi-turn setting by sequentially concatenating current questions, dialog history and schema items \cite{gou2020contextualize}. 

\textbf{Second}, the pre-trained language models (PLMs) such as BERT \cite{devlin2018bert} and RoBERTa \cite{liu2019roberta} can significantly boost parsing accuracy by enhancing the generalization of the encoder and capturing long-term token dependencies.
In general, for question tokens, the output hidden states from the final layer of the Transformer block in the BERT \cite{devlin2018bert} or RoBERTa models are considered as the contextualized representations of question tokens. For each database schema item, the output hidden state of its front special token ${\rm [SEP]}$ is regarded as the table or column header representation.

\textbf{Third}, leveraging flexible neural architecture on the top of PLMs can further enhance the encoder's output representations with strong expressive ability.
For example, SQLova \cite{hwang2019comprehensive} and SDSQL \cite{hui2021improving} further stack two Bi-LSTM layers on the top of output representations of BERT \cite{devlin2018bert}.
GAZP \cite{zhong2020grounded} proposes an additional self-attention layer \cite{vaswani2017attention} on the top of a Bi-LSTM layer to compute the intermediate representations.
RYANSQL \cite{choi2021ryansql} sequentially employs the convolutional neural network \cite{albawi2017understanding} with dense connection \cite{yoon2018dynamic} and a scaled dot-product attention layer \cite{vaswani2017attention} on the top of BERT \cite{devlin2018bert} to align question tokens with columns. It is noteworthy that the parameters of convolutional neural network are shared across the NL question and columns.   
BRIDGE \cite{lin2020bridging} encodes the input sequence with BERT \cite{devlin2018bert} and lightweight subsequent layers (i.e., two Bi-LSTM layers). In addition, dense look-up features are applied to represent meta-data information of the table schema such as primary key, foreign key and datatype. These meta-data features are further fused with the BERT \cite{devlin2018bert} encoding of the schema component via a feed-forward layer.

\subsubsection{Structure Modelling}
The development of large cross-domain datasets such as WikiSQL \cite{zhong2017seq2sql} and Spider \cite{yu2018spider} results in the realistic generalization challenge to deal with unseen table schemas. Each NL question corresponds to a multi-table database schema. The training and testing sets do not share overlapped databases.  The challenge of the generalization requires the text-to-SQL parsing methods to encode the NL question and the table schema into representations with powerful expressive ability from three aspects. 
\begin{itemize}
    \item First, the encoder should be able to recognize NL tokens used to refer to  tables and columns either implicitly or explicitly, which is called schema \textbf{linking structure} – aligning entity mentions in the NL question to the mentioned schema tables or columns. 
    \item Second, the encoded representations should be aware of the \textbf{schema structure} information such as the primary keys, the foreign keys, and the column types.
    \item Third, the encoder should be able to perceive complex variations in the NL question, i.e., \textbf{question structure}.
\end{itemize}

Graph are the best form of data to express the complex structure in the text-to-SQL parsing task.
Recently, several graph-based methods \cite{bogin2019representing,wang2019rat,cao2021lgesql} have been proposed to reason over the NL question tokens and schema entities, and model the complex input representation. 
These methods consider the NL question tokens and schema items as multi-typed nodes, and the structural relations (edges) among the nodes can be pre-defined to express diverse intra-schema relations, question-schema relations and intra-question relations. 

\begin{figure}
    \centering
    \includegraphics[width=8cm]{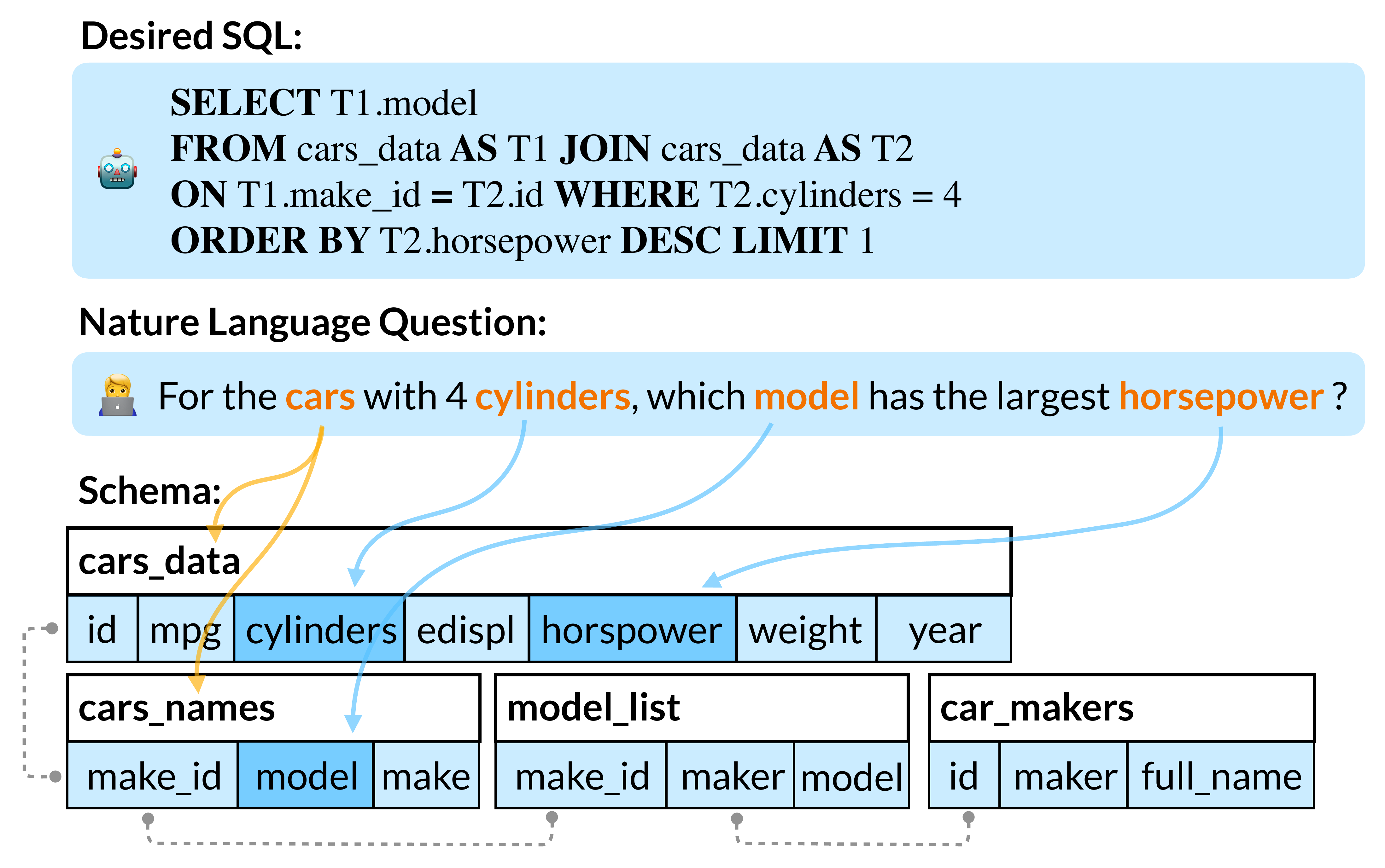}
    \caption{Example of schema linking structure used in \cite{wang2019rat}.}
    \label{linking}
\end{figure}

\vspace{0.3cm}
\noindent \textbf{Linking Structure} \quad
As illustrated in Figure \ref{linking}, schema linking aims at identifying references of columns, tables and condition values in NL questions \cite{liu2021awakening}. 
The text-to-SQL parsers should learn to detect table or column names mentioned in NL questions by matching question tokens with the schema, and the identified tables or columns are then utilized to generate SQL queries. Intuitively, schema linking facilitates both cross-domain generalizability and complex SQL generation, which have been regarded as the current bottleneck of text-to-SQL parsing. \cite{lei2020re} demonstrates that more accurate schema linking conclusively leads to better text-to-SQL parsing performance.
Conventional schema linking can be extracted by means of rules or string matching. For example, IRNET \cite{guo2019towards} takes the extracted linking information directly as input. SDSQL \cite{hui2021improving} leverages the extracted linking information as a label for multi-task learning.  IESQL \cite{ma2020mention} employs a conditional random filed (CRF) layer \cite{lafferty2001conditional} for the question segment, which transforms the linking task into a sequence tagging task. 
However, the aforementioned schema linking methods cannot capture the comprehensive semantic relationship between the NL question and table schema. 
Another popular approach \cite{zhang2019editing} proposes the cross-attention to implicitly learn relationships between the NL question and schema representations.
%Rule-based schema linking is weakly generalised, and another popular approach \cite{zhang2019editing} is used the cross-attention to implicitly learn links between natural language questions and schema representations.

Recently, the graph-based linking approaches \cite{wang2019rat, rubin2020smbop, huang2021relation, chen2021shadowgnn, cao2021lgesql} have been proposed to reason over the NL question tokens and schema entities, and model the complex input representation. 
These methods consider the NL question tokens and schema items as multi-typed nodes, and the structural relations (edges) among the nodes can be pre-defined to express diverse intra-schema relations, question-schema relations and intra-question relations. 
In particular, in RATSQL \cite{wang2019rat}, the graph is constructed based on two kinds of relations (i.e., name-based linking and value-based linking), where the name-based linking refers to partial or exact occurrences of table/column names in the same NL question and the value-based linking refers to the question-schema alignment that occurs when the NL question mentions any values appearing in the schema and the desired SQL. The relation-aware self-attention mechanism \cite{shaw2018self} is then proposed for graph representation learning, which exploits global reasoning over the constructed graph. Specifically, given a sequence of token representations, the relational-aware self-attention computes a scalar similarity score between each pair of token representations $e_{i j} \propto \mathbf{u}_{i} W_{Q}\left(\mathbf{u}_{j} W_{K}+\mathbf{r}_{i j}^{K}\right)$, where $\mathbf{u}_{i}$ and $\mathbf{u}_{j}$ denote the graph nodes, and the term $\mathbf{r}_{i j}^{K}$ denotes an embedding that represents a relation between $\mathbf{u}_{i}$ and $\mathbf{u}_{j}$ from a closed set of possible relations. To enhance the model generalization capability for unseen or rare schemas, ShadowGNN \cite{chen2021shadowgnn} alleviates the impact of the domain information by abstracting the representations of the NL question and the SQL query before applying the relation-aware graph computation \cite{wang2019rat}.
In addition, several works have been devoted to tackle the challenge of heterogeneous graph encoding for the text-to-SQL parsing. LGESQL \cite{cao2021lgesql} constructs an edge-centric graph from the node-centric graph as in RATSQL \cite{wang2019rat}, which explicitly considers the topological structure of edges. The information propagates more efficiently by considering both the connections between nodes and the topology of directed edges. Two relational graph attention networks (RGANs) \cite{wang2020relational} are devised to model the structure of the node-centric graph and the edge-centric graph respectively, mapping the input heterogeneous graph into token representations.

\vspace{0.3cm}
\noindent \textbf{Schema Structure} \quad 
It is intuitive to leverage a relational graph neural network (GNN) to model the relations in the relational database, propagating the node information to its neighbouring nodes. 
The GNN-based methods help to aggregate feature information of neighboring nodes, making the obtained input representation more powerful. 
Schema-GNN \cite{bogin2019representing} first converts the database schemas to a graph by adding three types of edges: the foreign-primary key relation, the column-in-table relation and the table-own-column relation. The constructed graph is softly pruned conditioned on the input question, which is then fed into gated GNNs \cite{li2015gated} to learn schema representations being aware of the global schema structure.
Furthermore, Global-GNN \cite{bogin2019global} proposes a similar approach by employing a graph convolutional network (GCN) to learn schema representations, where a relevance probability conditioned on the question is computed for every schema node. 
Some advanced studies, such as RAT-SQL~\cite{wang2019rat} and LGESQL \cite{cao2021lgesql}, also learn the structure of the schema as a unique edge in the graph, demonstrating the indispensability of the schema structure for text-to-SQL parsing.

\vspace{0.3cm}
\noindent \textbf{Question Structure} \quad
S$^2$SQL \cite{hui2022s} investigates the importance of syntax in text-to-SQL encoder, and proposes a flexible and robust injection method. It leverages three induct dependency types, i.e., \texttt{Forward}, \texttt{Backward}, \texttt{NONE}, which stack multi-layer transformers to implicitly model complex question structure. In addition, a decoupling constraint is employed to induce the diverse relation embedding.
SADGA \cite{cai2021sadga} constructs the question graph conditioned on the dependency structure and the contextual structure of the NL question sequence, and builds the schema graph conditioned on the schema structure. Specifically, three different types of links are defined for question tokens to construct the graphs: the 1-order word dependency (i.e., the relation between two consecutive
 words), the 2-order word dependency, and the parsing-based dependency that captures  syntactic relations among the NL question words. Then, the structure-aware aggregation approach is proposed to capture the alignment between the constructed graphs through two-stage linking.  The unified representations are learned by aggregating the information via a gated-based mechanism.

\subsection{Decoder}
The decoder used in existing Text-to-SQL parsing models can be divided into two categories: \textbf{sketch-based methods} and \textbf{generation-based methods}. In this section, we provide a comprehensive overview of the two types of decoder architectures.

\begin{figure}
    \centering
    \includegraphics[width=0.5\textwidth]{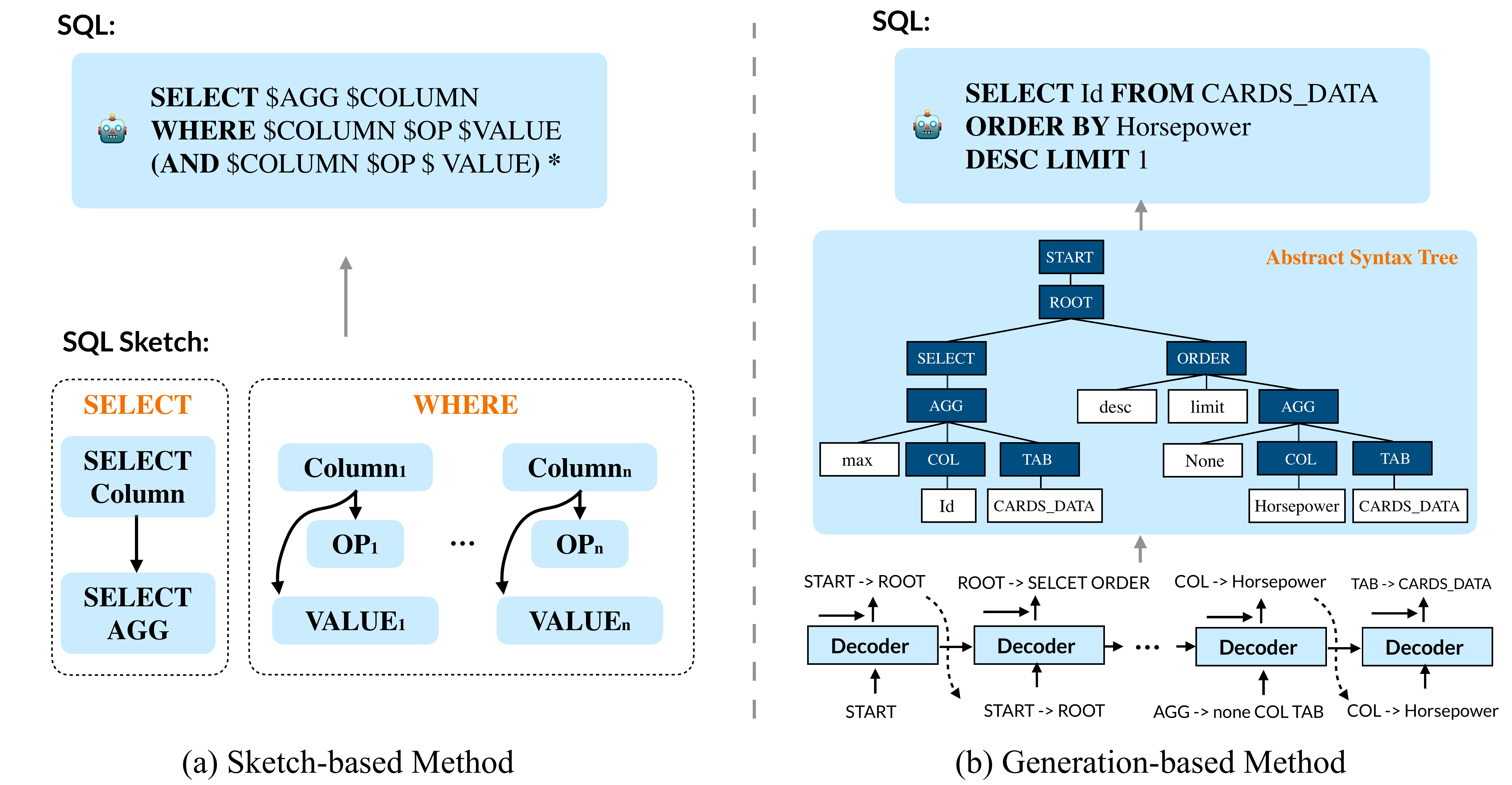}
    \caption{Example of SQL sketch used in \cite{xu2017sqlnet}. }
    \label{decoder}
\end{figure}

\subsubsection{Sketch-based Methods}
The sketch-based methods decompose the SQL generation procedure into sub-modules, e.g., \texttt{SELECT} column, \texttt{AGG} function, \texttt{WHERE} value. 
For example, SQLNet \cite{xu2017sqlnet} employs the SQL sketch. The tokens \texttt{SELECT}, \texttt{WHERE} and \texttt{AND} indicate the SQL keywords, and the following components indicate the types of prediction slots to be filled. For example, the \texttt{AGG} slot indicates the slot to be filled with either an empty token or one of the aggregation operators such as \texttt{SUM} and \texttt{MAX}; the \texttt{VALUE} slot needs to be filled with a sub-string of the question; the \texttt{COLUMN} slot needs to be filled with a column name; the \texttt{OP} slot needs to be filled with operations such as $>$, $<$, $=$). These slots are later gathered together and interpreted to generate the final SQL query. 
Each slot has separate model which does not share their trainable parameters and is responsible for predicting a part of the final SQL independently. Specifically, for the \texttt{COLUMN} slot, a column attention mechanism \cite{xu2017sqlnet} is applied to reflect the most relevant information in NL questions when prediction is made on a particular column. For the \texttt{OP} slot, predicting its value is a 3-way classification task ($>$, $<$, $=$). For the \texttt{VALUE} slot, \cite{xu2017sqlnet} employs a Seq2Seq structure to generate the sub-string of the NL question.

SQLova \cite{hwang2019comprehensive} and SDSQL \cite{hui2021improving} modify the  syntax-guided sketch used in \cite{xu2017sqlnet}. The proposed sketch-based decoder consists of six prediction modules, including  \texttt{WHERE-NUMBER}, \texttt{WHERE-COLUMN}, \texttt{SELECT-COLUMN}, \texttt{SELECT-AGGREGATION},   \texttt{WHERE-OPERATOR}, and \texttt{WHERE-VALUE}. The specific role of each action is described as follows:
\begin{itemize}
    \item \texttt{SELECT-COLUMN} identifies the column in ``\texttt{SELECT}'' clause from the given NL question. 
    \item \texttt{SELECT-AGGREGATION} identifies the aggregation operator for the given select-column prediction. 
    \item \texttt{WHERE-NUMBER} predicts the number of ``\texttt{WHERE}'' conditions in SQL queries. 
    \item \texttt{WHERE-COLUMN} calculates the probability of generating each columns for the given NL question.
    \item \texttt{WHERE-OPERATOR} identifies the most probable operators given where-column prediction among three possible choices ($>$, =, $<$). 
    \item \texttt{WHERE-VALUE} identifies which tokens of a NL question correspond to condition values for the given ``\texttt{WHERE}'' columns.
\end{itemize}
In the SQL query generation stage,  an execution-guided decoding strategy \cite{wang2018execution} is utilized to exclude the non-executable partial SQL queries from the output candidates.
TypeSQL \cite{yu2018typesql} further improves the above approach by declining the number of modules. TypeSQL chooses to combine the select-column module and the where-column module into a single module since their prediction procedures are similar, and the where-column module depends on the output of the select-column module. In addition,  the where-operator and where-value modules are combined together because the predictions of these two modules depend on the outputs of the where-column module.
Generally, the sketch-based approaches are fast and guaranteed to conform to correct SQL syntax rules. However, it is difficult for these approaches to handle complex SQL statements such as multi-table JOINs, nested queries, and so on. 
Thus, the sketch-based approaches are popular on the WikiSQL \cite{zhong2017seq2sql} dataset, but are difficult to be applied on the Spider \cite{yu2018spider} dataset which involves complex SQLs. Only RYANSQL \cite{choi2021ryansql} implements complex SQL generation by recursively applying the sketch method.

\subsubsection{Generation-based Methods}
On the other hand, the generation-based approaches are based on the Seq2Seq model to decode SQL, which are more preferable for complex SQL scenarios than the sketch-based approaches. For example, Bridge \cite{lin2020bridging} uses a LSTM-based pointer-generator \cite{see2017get} with multi-head attention and copy mechanism as the decoder which is initiated with the final state of the encoder. At each decoding step, the decoder performs one of the following actions: generating a token from the vocabulary $V$, copying a token from the question $Q$, or copying a schema component from the database schema $S$. 

Since the above generation-based approaches may not generate SQL queries with correct grammar, 
some advanced methods \cite{guo2019towards,wang2019rat} generate the SQL as an abstract syntax tree (AST) \cite{wang1997zephyr} in the depth-first traversal order \cite{tarjan1972depth}. In particular, these methods employ an LSTM decoder to perform a sequence of three types of actions that either expand the last generated node into a grammar rule, called \texttt{APPLY-RULE} action or choose a column/table from the schema when completing a leaf node, called \texttt{SELECT-COLUMN} action and \texttt{SELECT-TABLE} action respectively. These actions can construct the corresponding AST of the target SQL query. 
Specifically, \texttt{APPLY-RULE} applies a production rule to the current derivation tree of a SQL query and expands the last generated node with the grammar rule. The probability distribution is computed by a softmax classification layer for the pre-defined abstract syntax description language (ASDL) rules. 
\texttt{SELECT-COLUMN} and \texttt{SELECT-TABLE} complete a leaf node by selecting a column $c$ or a table $t$ from the database schema respectively by directly copying the table and column names from database schema via copy mechanism \cite{see2017get}.

There are also several works \cite{xu2020optimizing,shaw-etal-2021-compositional,xuan2021sead} which neglect the SQL grammar during the decoding process, by leveraging the powerful large scale pre-trained language model like T5 \cite{raffel2019exploring} finetuned on the text-to-SQL training set for SQL query generation. Formally, the transformer-based decoder follows the standard text generation process and produces the hidden state in step $t$ for generating the $t$-th token as described in \cite{vaswani2017attention}. An affine transformation is then applied on the learned hidden state to obtain prediction probability over the target vocabulary $V$ for each word. 
In addition, as revealed in \cite{kelkar2020bertrand}, for some cases, although the best generated SQL is in the candidate list of beam search, it is not at the top of the candidate list. Therefore, a discriminative re-ranker (\textbf{Re-Ranking}) strategy is introduced to extract the best SQL query from the candidate list predicted by the text-to-SQL parser. The re-ranker is constructed as a BERT fine-tuned classifier that is independent of schema, and  the probability of the classifier is utilized as the score for the query to re-rank. 
Some studies \cite{guo2019towards, chen2021shadowgnn} introduce a domain specific language (\textbf{DSL}) serving as an intermediate representation to bridge  the NL question and the SQL query.
The methods reveal that there is a invariable mismatch between the intentions conveyed in natural language and the implementation details in SQL when the resulting SQL is processed to tree-structured form.
Hence, the key idea of DSL is to omit the implementation details of the intermediate representations. 
IRNet \cite{guo2019towards} presents a grammar-based neural model to generate a SemQL query as the
intermediate representation bridging the NL question and the SQL query, and a SQL query is then inferred from the generated SemQL query with domain knowledge.
%From the synthesized DSL query with domain knowledge, IRNet finally infers a SQL query.
Furthermore, a \textbf{constrained decoding} strategy is proposed in Picard \cite{scholak2021picard} via incremental parsing, which facilitates the parser to identify valid target sequences by rejecting inadmissible tokens at each decoding step.
%Picard \cite{scholak2021picard} further proposes \textbf{constrained decoding} of language models via incremental parsing and helps to find valid output sequences by rejecting inadmissible tokens at each decoding step.

\section{Multi-turn T2S Parsing Approaches}
Compared to the single-turn T2S setting, the multi-turn T2S setting emphasizes the usage of contextual information (historical information), which can be incorporated in both the encoder and the decoder. Next, we describe how the contextual information is leveraged.

\subsection{Encoder}
The encoder processes the contextual information for input representation learning. In addition, the linking structure and the schema structure are considered during the encoding phase. 
%For encoders, firstly contextual information for  needs to be considered during the \textbf{input representation}, secondly evolution is necessary when modelling structures, especially linking structures and schema structures.

\begin{figure}
    \centering
    \includegraphics[width=0.5\textwidth]{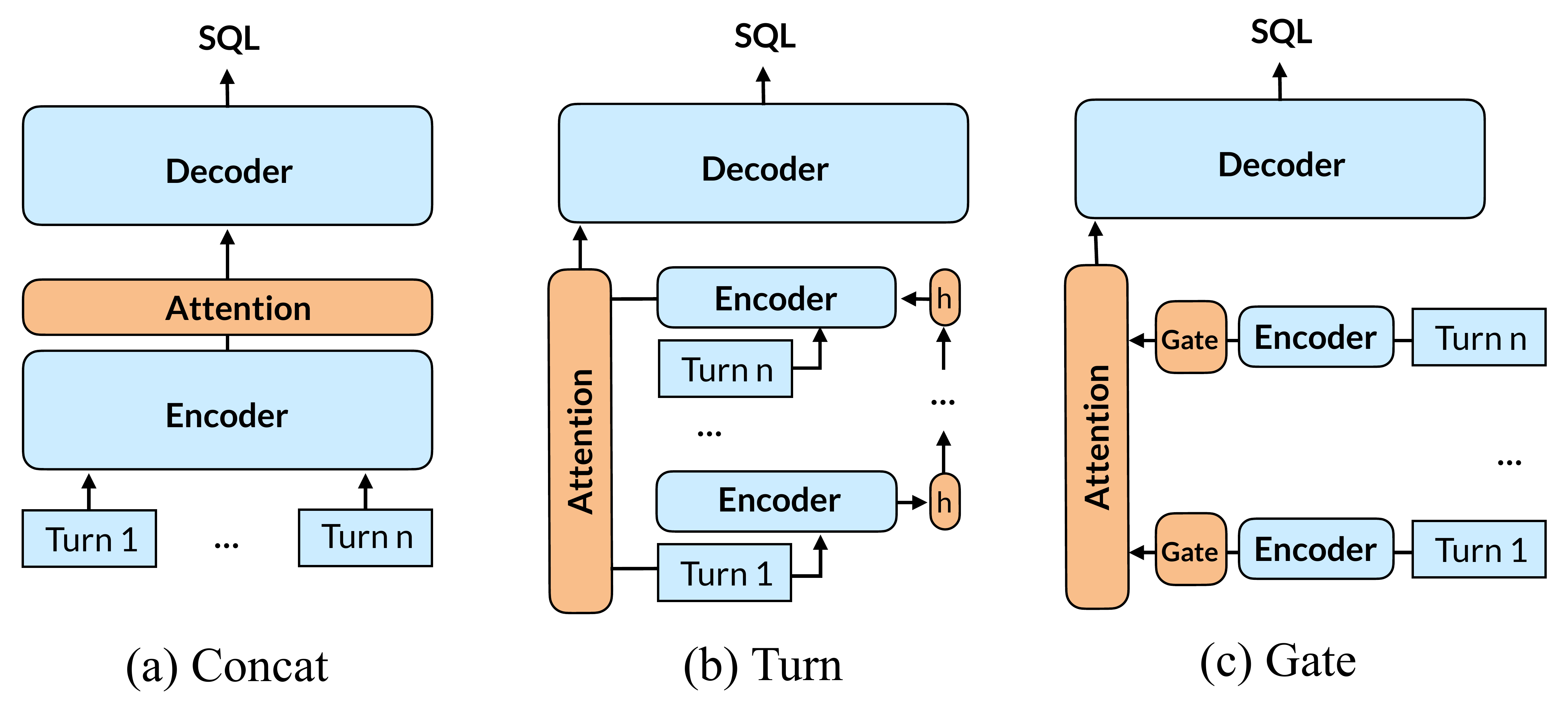}
    \caption{Different contextual NL questions encoding strategies for multi-turn T2S parsing \cite{Liu2020HowFA}. }
    \label{context}
\end{figure}

\subsubsection{Multi-turn Input Representation}
With regard to the multi-turn representation learning, previous works mainly focus on two aspects, including (i) how to learn high-quality contextual question and schema representations and (ii) how to effectively encode the historical SQL queries.

For the contextual question and schema representation learning, as shown in Fig. \ref{context}, \cite{Liu2020HowFA} investigates the impact of different contextual information encoding methods on the multi-turn T2S parsing performance, including (a) contacting all the NL questions within each question sequence as input, (b) using a turn-level encoder to deal with each question, and (c) devising a gate mechanism to balance the importance of each historical question. 
Typically, EditSQL \cite{zhang2019editing} utilizes two separate Bi-LSTMs \cite{hochreiter1997long} for encoding the NL questions and the table schema respectively. Specifically, for the question at each turn, EditSQL first utilizes a Bi-LSTM to encode the question tokens. The output hidden states are then fed into a dot-product attention layer \cite{luong2015effective} over the column header embeddings. The relationships among the multi-turn questions are captured by the turn attention mechanism. At the current turn,  the dot-product attention between the current question and previous questions in the history is computed, and the weighted average of previous question representations is then added to the current question representation to form the context-aware question representation. 
For each column header, the table name and the column name are concatenated and passed into a Bi-LSTM layer. The output hidden states are then fed into a self-attention layer \cite{vaswani2017attention} to better capture the internal structure of the table schemas such as foreign keys. In addition, the self-attention vector and the question attention vector are concatenated and fed into the second Bi-LSTM layer to obtain the final column header representation.
\citep{memsp} learns a context memory controller to maintain the memory by keeping the cumulative meaning of the sequential NL questions and using an external memory to represent contextual information.
\citep{chen2021decoupled} decouples the multi-turn T2S parsing into two pipeline tasks: question rewriting and single-turn T2S parsing. The question rewriting (QR) module aims to generate semantic-completion question based on the dialogue context, which concatenates the dialogue history and the latest question as the input of the QR module. The goal of the QR module is to generate a simplified expression based on the latest question and the dialogue history. The single-turn T2S parser predicts the corresponding SQL query with the simplified expression generated by the QR module, which chooses a pre-trained BART as question rewriter and  RAT-SQL \cite{wang2019rat} as the single-turn text-to-SQL parser.

For the historical SQL input encoding, HIE-SQL \cite{zheng2022hie} treats the logic-forms of the SQL query as an another modality for the NL question and incorporates additional SQL encoders to capture the semantic connection between the SQL query and the NL question. It produces comprehensive representations of the previously predicted SQL queries to improve the contextual representation.

\subsubsection{Multi-turn Structure Modelling}
Different from the single-turn setting, the multi-turn setting requires the integration of contextual inductive bias into the structural modelling. 

\vspace{0.3cm}
\noindent \textbf{Linking Structure} \quad 
R$^2$SQL \cite{hui2021dynamic} focuses on the uniqueness of contextual linking structures, introducing a novel dynamic graph framework to efficiently model contextual questions, database schemas, and the complex linking structures between them.  A decay mechanism is applied to mitigate the impact of historical links on the current turn. 

\vspace{0.3cm}
\noindent \textbf{Schema Structure} \quad
% IGSQL \cite{cai2020igsql} proposed a schema interaction graph encoder to model context consistency of historical schema representations. First, a database schema graph was constructed by taking schema entities as graph nodes and primary-foreign keys as graph edges.  Then, a graph encoder, which is composed of cross-turn schema interaction graph layers for updating schema item representations with previous question and intra-turn schema graph layers for aggregating adjacent item representations with the same question, was proposed to capture the complex semantic information of text-to-SQL data. 
% IST-SQL \cite{wang2020tracking} focused on modeling intra-schema relations and defined two types of relations (foreign-key relation and foreign-key-table relation) between two schema items. For the foreign-key relation, we construct an edge with  ``IN'' or ``OUT'' directions between two column names if there is a foreign key pair in the database. For the foreign-key-table relation, we build an edge with  ``IN'' or ``OUT'' directions between two column names if the two columns are in different tables that have one or more foreign key pairs.  After constructing the graph, the relational graph neural network (RGNN) is proposed to propagate node information to nearby nodes and calculate schema representations.
\citep{zhang2019editing} exploits the conversation history by editing the previous predicted SQL to improve the generation quality.
It focuses on taking advantages of previous question texts and previously predicted SQL query to predict the SQL query at the current turn.
Concretely, the previous question is first encoded into a series of tokens, and the decoder then predicts a switch to alter it at the token level. 
This sequence editing approach simulates the changes at token level and is hence resistant to error propagation.
IGSQL \cite{cai2020igsql} points out that it should not only take historical user inputs and previously predicted SQL query into consideration but also utilize the historical information of database schema items.
Therefore, IGSQL proposes a database schema interaction graph encoder to learn database schema items together with historical items, keeping context consistency for context-dependent text-to-SQL parsing.
The cross-turn schema interaction graph layer and intra-turn schema graph layer update the schema item representations by using  the previous turn and the current turn respectively.
IST-SQL \cite{wang2021tracking} deals with the multi-turn text-to-SQL parsing task inspired by the task-oriented dialogue generation task.
IST-SQL defines, tracks and utilizes the interaction states for multi-turn text-to-SQL parsing, where each interaction state is updated by a state update mechanism based on the previously predicted SQL query.

\subsection{Decoder}
For the multi-turn setting, most previous methods \cite{zhang2019editing, wang2020tracking, hui2021dynamic} employ a LSTM decoder with attention mechanisms to produce SQL queries conditioned on the  historical NL questions, the current NL question, and the table schema. 
The decoder takes the encoded representations of the current NL question, SQL-states, schema-states, and last predicted SQL query as input and apply query editing mechanism \cite{zhang2019editing} in the decoding progress to edit the previously generated SQL query while incorporating the context of the NL questions and schemas. To further alleviate the challenge that the tokens from the vocabulary may be completely 
irrelevant  to the SQL query, separate layers are used to predict SQL keywords,  table names and question tokens.  A softmax operation is finally used to generate the output probability distribution. 

\section{Pre-training for Text-to-SQL Parsing}
Pre-trained language models (PLMs) have proved to be powerful in enhancing text-to-SQL parsing and yield impressive performances, which benefit from the rich  knowledge in large-scale corpus. However, as revealed in previous works \cite{yin2020tabert,yu2020grappa}, there are intrinsic differences between the distribution of tables and plain texts, leading to sub-optimal performances of general PLMs such as BERT \cite{devlin2018bert} in text-to-SQL parsing.
Recently, several studies have been proposed to alleviate the above limitation and build tabular language models (TaLMs) by simultaneously encoding tables and texts, which show improved results on downstream text-to-SQL parsing tasks.
In this section, we provide a comprehensive review of existing studies on pre-training for text-to-SQL parsing from the perspectives of pre-training data construction, input feature learning, pre-training objectives and the backbone model architectures.

\subsection{Pre-training Data Construction}
Insufficient training data is an important challenge to learn powerful pre-trained tabular language models. The quality, quantity and diversity of the pre-training data have significant influence on the general performance of the pre-trained language models when applied to downstream text-to-SQL parsing tasks. Although it is easy to collect a large amount of tables from Web (e.g., Wikipedia), obtaining high-quality NL questions and their corresponding SQL queries over the collected tables is a labor-intensive and time-consuming process.  
Recently, there have been plenty of studies to generate pre-training data for text-to-SQL parsing manually or automatically. 
Next, we discuss the previous pre-training data construction methods from three perspectives: table collection, NL question generation, and logic form (SQL) generation.

% Please add the following required packages to your document preamble:
% \usepackage{booktabs}

\begin{table*}[!t]
\centering
\caption{The pre-training data construction.}
\resizebox{\textwidth}{!}{%
\begin{tabular}{@{}l|cccccccc|ccc@{}}

\toprule
          & \multicolumn{8}{c|}{Table}                                 & \multicolumn{2}{c}{Question} \\
          & WIKITABLEQUESTIONS & WIKISQL \cite{zhong2017seq2sql} & WIKITABLE & ToTTo & Spider \cite{yu2018spider} &WebTable &Wikipedia & GITHUB & ToTTo & Wikipedia & Spider \cite{yu2018spider}\\
\midrule
TaBERT\cite{yin2020tabert}    &                    &         &           &                &   & \ding{51}  & \ding{51} && && \\
\textsc{TaPas} \cite{herzig-etal-2020-tapas}     &                    &         &           &                  &   &   &\ding{51} && & \ding{51}& \\
\textsc{Grappa} \cite{yu2020grappa}    &                    &\ding{51}   & \ding{51}   &       & \ding{51}          &  &  && && \\
GAP         &                    &  &    &       &          &  &  & \ding{51} & &&\ding{51} \\
\textsc{StruG}     &                    &         &           &\ding{51}                &   &  &  & &\ding{51}  & &\\
MATE     &                    &         &           &                  &   &   &\ding{51} && & \ding{51}& \\
TAPEX     &  \ding{51}        &         &           &                  &   &   & && && \\
TableFormer    &                    &         &           &                  &   &   &\ding{51} && & \ding{51}& \\
SCORE     &                    &\ding{51}     &\ding{51}      &       & \ding{51}      &  & && &\ding{51} & \\
% H-NEURSYN &                    &         &           &       &        &               &     \\
\bottomrule
\end{tabular}
}
\end{table*}

\subsubsection{Table Collection}
We briefly introduce several sources that have been extensively used for table collection. 
WikiTableQuestion \cite{pasupat2015compositional} is a representative corpus which is composed of 22,033 question-answer pairs on 2,108 tables, where the tables are randomly collected from Wikipedia with at least five columns and eight rows. The tables are pre-processed by omitting all the non-textual information, and each merged cell is duplicated to keep the table valid. In total, there are 3,929 distinct column headers (relations) among the 13,396 columns.
WikiSQL \cite{zhong2017seq2sql} is a collection of 80,654 hand-crafted question-SQL pairs along with 24,241 HTML tables collected from Wikipedia. The tables are collected from \cite{bhagavatula2013methods}, and the small tables that have less than five columns or five rows are filtered. 
WDC WebTables \cite{lehmberg2016large} is a large-scale table collection, which contains over 233 million tables and has been extracted from the July 2015 version of the CommonCrawl. Those tables are classified as either relational (90 million), entity (139 million), or matrix (3 million). 
WikiTables  \cite{bhagavatula2015tabel} contains 1.6 million high-quality relational Wikipedia tables constructed by extracting all HTML tables from Wikipedia which had the class attribute ``wikitable'' (used to easily identify data tables) from the November 2013 XML dump of English Wikipedia.
ToTTo \cite{parikh2020totto} is an open-domain English table-to-text dataset with over 120,000 training examples which contains 83,141 corresponding web tables automatically collected from Wikipedia using heuristics with various schema spanning various topical categories.
%FUSE \yangmin{Pls add descriptions.}
%TableArXiv \yangmin{Pls add descriptions.}
WebQueryTable \cite{sun2019content} is composed of 273,816 tables by using 21,113 web queries, where each query is used to search the Web pages and the relevant tables are obtained from the top ranked Web pages.  
Spider \cite{yu2018spider} is a large-scale text-to-SQL dataset, which contains 200 databases belonging to 138 different domains. In particular, the databases come from three resources: (i) 70 complex databases are collected from SQL tutorials, college database courses, and textbook examples, (ii) 40 databases are collected from the DatabaseAnswers\footnote{http://www.databaseanswers.org/}, (iii) 90 databases are selected from WikiSQL \cite{zhong2017seq2sql}.

% TUTA \cite{wang2021tuta} collects about 53.4M web tables from WikiTable\footnote{https://github.com/bfetahu/wiki\_tables} as well as the WDC WebTable dataset \cite{lehmberg2016large} and crawls about 13.5M spreadsheet tables from different web sites. 
% After performing data pre-processing and filtering, the final created corpus contains about 57.9M tables in total. 

%Since it is easy to obtain high-quality large-scale tables from web, in this section, we briefly introduce several sources that have been extensively used for table collection. \cite{liu2021tapex} employs tables in WikiTableQuestion \cite{pasupat2015compositional}, which is a dataset containing 22,033 pairs of questions and answers based on 2,108 Wikipedia tables. \cite{yu2020score} use about 400,000 tables in WikiTable \cite{bhagavatula2015tabel} and WikiSQL \cite{zhong2017seq2sql}. \cite{yu2020grappa} utilize about 166 tables in Spider dataset \cite{yu2018spider}. \cite{deng2020structure} employs an existing large-scale table-to-text generation dataset called ToTTo \cite{parikh2020totto} which contains 83,141 corresponding web tables automatically collected from Wikipedia with various schema spanning various topical categories.

\subsubsection{Natural Language Question Annotation}
So far, different question annotation methods have been introduced to annotate natural language questions based on the collected databases. Generally, the methods can be divided into three categories: sampling-based methods, template-based methods, and generation-based methods. 

\vspace{0.3cm}
\noindent \textbf{Sampling-based Methods~~}
Many works produce the NL questions in pre-training data by extracting text-table pairs from Wikipedia.
Concretely, \textsc{TaPas} \cite{herzig-etal-2020-tapas} creates the pre-training corpus by collecting text-table pairs from Wikipedia, where there are about 6.2M tables and 21.3M text snippets. In particular, the table captions, segment titles, article descriptions, article titles, and textual table segments are extracted as the text snippets of the corresponding tables.
\textsc{StruG} \cite{deng2020structure} directly collects about 120k NL web tables and corresponding text descriptions from the ToTTo dataset \cite{parikh2020totto}, which extracts the text-table pairs from Wikipedia by employing three heuristics: number matching, cell matching, and hyperlinks.

\vspace{0.3cm}
\noindent \textbf{Template-based Methods~~}
There are several works that generate the NL questions automatically by using templates or rules.
%\textsc{GAZP} \cite{zhong2020grounded} 
\textsc{Grappa} \cite{yu2020grappa} constructs question-SQL templates by extracting entity mentions of SQL operations and database schemas. By leveraging the created templates on randomly sampled tables, a large amount of question-SQL pairs can be synthesized automatically.  
\textsc{SCoRe} \cite{yu2020score} leverages only  500  samples from the development set of \textsc{SParC} \cite{yu2019sparc}  to derive utterance-SQL generation grammars consisting of a list of synchronous question-SQL templates and follow-up question
templates. Finally, about 435k text-to-SQL conversations are synthesized for context-dependent text-to-SQL pre-training. 

\vspace{0.3cm}
\noindent \textbf{Generation-based Methods~~}
Several works have been introduced to generate NL questions from entity sequences automatically with text generation models. For example, \cite{yang2021hierarchical} proposes a cross-domain neural model, which accepts a table schema and samples a sequence of entities to be appear in the NL question, to transform the entity sequence to the NL question. Specifically, the T5 model \cite{raffel2019exploring} is first fine-tuned on a small corpus containing entity-question pairs and then applied to generate NL questions given entity sequences. For example, given the input entity sequence ``\textit{department management : head name text} | \textit{head age number} | \textit{head born state text}'', the T5 model is likely to output the NL question ``\textit{List the name, born state and age of the heads of departments ordered by age.}''.
In addition, \cite{wang2021learning} introduces a generative model to generate utterance-SQL pairs, which leverages the probabilistic context-free grammar (PCFG) to model the SQL queries and then employs a BART-based translation model to transform the logical forms to NL questions. For example, for a given input SQL sequence ``\textit{select area where state\_name = ‘texas’}'', the generative model outputs the NL question ``\textit{what is the area of Texas?}''.
GAZP \cite{zhong2020grounded} also generates utterances corresponding to these logical forms using the generative model. 
In addition, the input and output consistency of the synthesized utterance is verified. Specifically, we parse the generated queries into logical forms and keep the queries whose parses are equivalent to the corresponding original logical forms. 
%, and we keep the utterances  keeping those whose parses are equivalent to the original sampled logical form.

There are also some studies that invite annotators to manually create natural questions and corresponding SQL queries without leveraging templates or rules, such that the generated NL questions are natural and diverse. In particular, Spider \cite{yu2018spider} generates 10,181 NL questions and 5,693 unique SQL queries over 200 databases by considering three primary aspects: SQL pattern coverage ensuring that enough SQL samples are obtained to cover all common SQL patterns, SQL consistency ensuring that the semantically equivalent NL questions share the same SQL query, and question clarity ensuring that the vague or too ambiguous questions are not included.

\subsubsection{SQL Annotation}
Generally, the SQL annotation methods can be divided into three primary categories: logic perturbation, SQL template instantiation and hierarchical synthesis pipeline.

\vspace{0.3cm}
\noindent \textbf{Logic Perturbation~~}
Due to the expensive process of obtaining SQL queries, the logic perturbation-based approaches have been proposed to augment the SQL queries by performing random logic perturbation according to hand-tuned rules \cite{shu2021logic}. 
In particular, \cite{shu2021logic} generally enumerates the perturbations of each given SQL query based on hand-tuned rules that follow three kinds of logic inconsistencies: (i) logic shift aiming to generate the questions and logical forms that are logically distinct from the original ones, (ii) the phrase and number changes aiming to modify the appointed numerical values and phrases in logical forms, and (iii) entity insertion,  swapping and deletion that ignores the entity mention in logical forms, inserts
new entities into logical forms, or swaps any two entities within in a logical form.
There are three reasons for automatically generating more SQL queries by perturbing the logical forms. First, the regular structures of logical forms make the procedure of logical corruption controllable. Second, we can easily validate the perturbed logical forms with corresponding grammar checker and parser. Third, it is easy to obtain the corresponding questions of the generated SQL queries with minor modification given the original question-SQL pair.

\vspace{0.3cm}
\noindent \textbf{SQL Template Instantiation~~}
There are several studies that apply the SQL template instantiation methods to automatically generate SQL queries based on existing templates \cite{liu2021tapex, zhong2020grounded} or self-defined synchronous context-free grammar (SCFG) \cite{yu2020grappa}. \cite{liu2021tapex} utilizes the production rules of the SQL grammar defined in the SUQALL dataset \cite{shi2020potential} for SQL annotation. Given a SQL template, the headers and cell values of the tables are uniformly select to fill the template. 
GAZP \cite{zhong2020grounded} samples logical forms by leveraging a grammar, such as the SQL grammar over the database schema.
First, GAZP creates coarse templates by pre-processing the SQLs via the SQL grammar and further replacing the mentions of columns with typed slots.
Then, the slots of each coarse templates are filled-in by new database contents.
Instead of completely relying on the existing templates, \textsc{Grappa} \cite{yu2020grappa} learns from the examples in Spider \cite{yu2018spider} and designs a new SCFG which is then applied on a large number of existing tables to produce new SQL queries. The key idea behind this method is to define a set of non-terminal types for operations, table names, cell values, column names, and then substitute the entities with corresponding non-terminal symbols in the SQL query to form a SQL production rule. The SQL template instantiation methods often heavily depend on limited templates, and it is hard for them to generate diverse SQL queries with new compositions.

\vspace{0.3cm}
\noindent \textbf{Hierarchical Synthesis Pipeline~~}
Different from the above mentioned approaches that synthesize new SQL queries based on hand-crafted or induced rules and templates, the hierarchical synthesis pipeline approaches are based on the large-scale pre-trained language models (PLMs) \cite{raffel2019exploring}, which are motivated by the fact that PLMs can improve model generalization by incorporating additional diverse samples into the training corpus without labour-intensive manual works. Concretely, \cite{yang2021hierarchical} proposes a neural approach without grammar engineering but achieves high semantic parsing performance. 
To this end, the pre-trained text generation models such as T5 \cite{raffel2019exploring}, which is fine-tuned on the text-to-SQL data, are implemented to map entity sequences sampled from the table schema to NL questions \cite{yang2021hierarchical}. Then, the learned semantic parsers are then applied on the generated NL questions to produce the corresponding SQL queries. The overall data synthesis pipeline is easy to implement and achieves great diversity and coverage due to the usage of the large PLMs.

\subsection{Input Encoding}
In the text-to-SQL parsing task, the input often involves two parts: NL questions and table schemas, and the output would be the SQL queries. However, the textual data, tabular data and SQL queries are heterogeneous, which have different structures and formats. To be specific, the tabular data is generally distributed in two-dimensional structures with numerical values and words, while the SQL queries are usually composed of SQL keywords (e.g., ``SELECT'', ``UPDATE'', ``DELETE'', ``INSERT INTO'') and schema entities. Hence, it is non-trivial to develop a joint reasoning framework over the three types of data. In this section, we review the recent studies on heterogeneous input encoding of pre-training for text-to-SQL parsing.

%Due to the similarity between them, it is convenient to leverage language model to encode them together, which both treat them as the collection of tokens. However, text, table and SQL data are heterogeneous and actually have different input structures and formats. Tabular data is distributed in a semi-structured way, which has 2-D structure, including vertical columns (identifiable by column names with s specific number) and horizontal rows (with an unrestricted number). SQLs, logical form and other kinds of programming languages is usually composed of key words, i.e. select, where, def, etc, numerical values and operators, i.e. <, <=. In this way, works on table pre-training need to consider different encoding paradigms.

\subsubsection{Textual Data Encoding}
% Similar to general text encoding in various NLP tasks, the textual data encoding techniques for text-to-SQL parsing can be divided into static and dynamic approaches. 
% Some methods \cite{} \yangmin{pls add more suitable citations rather than RATSQL \cite{wang2019rat} and LGESQL \cite{cao2021lgesql} that also adopt PLMs.} applied GloVe \cite{pennington2014glove} to initialize the word embedding of each input tokens by looking up the embedding dictionary. 
Text encoding can be divided into dynamic and static types based on the word encoding in natural language processing. 
Some methods applied GloVe \cite{pennington2014glove} to initialize the word embedding of each input item by looking up an embedding dictionary without the context such as RATSQL \cite{wang2019rat} and LGESQL \cite{cao2021lgesql}.
However, the static embedding methods are still limited.
Neither of the static methods are able to tackle the  polysemy problem. In addition, the learned features are restricted by the pre-defined window size. With the development of the pre-train language models, some studies attempt to encode textual data with PLMs instead of static word embeddings. 
%Many of them are tokenized using WordPiece, encoded via the token vocabulary.
In particular,  plenty of methods (e.g., \textsc{TaBERT} \cite{yin2020tabert}, \textsc{TaPas} \cite{herzig-etal-2020-tapas}, MATE \cite{eisenschlos2021mate}, \textsc{StruG} \cite{deng2020structure}) utilize the pre-trained BERT \cite{devlin2018bert} as the encoder to get the contextualized word-level representations, and the parameters of BERT \cite{devlin2018bert} are updated along with the training process. \textsc{Grappa} \cite{yu2020grappa} uses \textsc{RoBerta} \cite{liu2019roberta} as the encoder. \textsc{Tapex} \cite{liu2021tapex} leverages both BART \cite{lewis-etal-2020-bart} encoder and decoder, while GAP \cite{shi2021learning} merely utilizes the BART encoder. 
%They all utilize BPE tokenization and the usage of PLMs is same as BERT \cite{devlin2018bert}.

\subsubsection{Tabular Data Encoding}
Different from textual data, the tabular data is distributed in two-dimensional (2-D) structures. The table pre-training approaches need to first convert the 2-D table data into linearized 1-D sequence input before feeding the tabular data into language models. A common serialization method is to flatten the table data into a sequence of tokens in the row-by-row manner and then concatenate the question tokens before the table tokens for tabular pre-training, such as \textsc{TaPas} \cite{herzig-etal-2020-tapas}, \textsc{MATE} \cite{eisenschlos2021mate}, and \textsc{TableFormer} \cite{yang2022tableformer}. \textsc{TaBERT} \cite{yin2020tabert} proposes content snapshots to encode a subset of table content which is most relevant to the input utterance. This strategy is then  combined with a vertical attention mechanism, sharing the information among the cell representations in different rows. 
There are also some studies (e.g., \textsc{StruG} \cite{deng2020structure}, \textsc{GraPPa} \cite{yu2020grappa} and UnifiedSKG \cite{xie2022unifiedskg}) which only take the headers of tables as input without considering the data cells. 

While the NLP models generally take the 1-D sequences as input, the positional encoding becomes crucial for tabular data to facilitate the neural models to better capture the structure information.
Most previous pre-training methods such as \textsc{TaBERT} \cite{yin2020tabert}, \textsc{GraPPa} \cite{yu2020grappa}, and \textsc{TaPEx} \cite{liu2021tapex}  explored the global positional encoding strategy on the flattened tabular sequences.
Nevertheless, in addition to the 1-D sequential positions, the tables have structured  columns and rows which consists of critical two-dimensional and hierarchical information.
The works such as \textsc{TaPas} \cite{herzig-etal-2020-tapas} and \textsc{MATE} \cite{eisenschlos2021mate} encode the row and column content based on column/row IDs.
\textsc{TableFormer} \cite{yang2022tableformer} decides whether two cells are in the same column/row and the column header, rather than considers the absolute order information of columns and rows in the tables. 

%In general, works on table pre-training need to first obtain linearized tabular sequences before feeding them into language models, since now language models only take 1-D sequence input. A common serialization method is to linearize raw tables row by row. Most works such as \textsc{Tapas} \cite{herzig-etal-2020-tapas}, \textsc{MATE} \cite{eisenschlos2021mate}, and \textsc{TableFormer} \cite{yang2022tableformer}. TaBERT \cite{yin2020tabert} only linearizes most relevant rows to the input utterance, and STRUG \cite{deng2020structure}, GraPPa \cite{yu2020grappa} and UnifiedSKG \cite{xie2022unifiedskg} only take headers as input without data cells.

%As mentioned above, tabular data is distributed in a semi-structured way, which has 2-D structure. While the NLP models now only take 1-D input sequence, the positional encoding becomes crucial  in tabular data to help model understand the table structure. By default, based on BERT, some works used global positional encoding in the flattened tabular sequences such as TaBERT, TaPas, MATE, STRUG, GraPPa, TaPEx and UnifiedSKG. However, in addition to 1D sequential positions, tables have structured rows and columns that contain critical two-dimensional and hierarchical information, and they are also desirable to be explicitly or implicitly encoded. Works such as Tapas and MATE learn column and row encodings based on column/row IDs. TableFormer captures the structural information with a same column/row relation as one kind of relative position between two linearized tokens.

% \subsection{Model Architecture}

\begin{table*}[!t]
\centering
\caption{The pre-training objectives for text-to-SQL parsing.}

\begin{tabular}{lllllll}
\toprule
% \multirow{2}{*}{\textbf{Models}}       &
% \multicolumn{5}{c}{Pretraining Objective}                                                           \\ \cline{2-6}
%                                 & Masked Language Modelling & Schema Linking & SQL Executor & Text Generation & Turn Contextual Swith \\
% \textbf{Models} 
\diagbox{\textbf{Models}}{\textbf{Objectives}} & \textbf{Masked Language Modelling} & \textbf{Schema Linking} & \textbf{SQL Executor} & \textbf{Text Generation} & \textbf{Context Modelling} \\
\midrule
TaBERT \cite{yin2020tabert}     & \ding{51} (MLM, MCP, CVR)   &                &              &                 &                       \\
\textsc{TaPas}  \cite{herzig-etal-2020-tapas}      & \ding{51} (MLM)          &                &              &                 &                       \\
\textsc{Grappa} \cite{yu2020grappa}     & \ding{51} (MLM)       &\ding{51} (SSP)     &              &                 &                       \\
GAP    \cite{shi2021learning}     & \ding{51} (MLM, CRec)  &\ding{51} (CPred)     &              &\ding{51} (GenSQL)      &                  \\
\textsc{StruG} \cite{deng2020structure}   \cite{}     &                         &\ding{51} (CG, VG)     &              &                 &                       \\
MATE \cite{wang2020meta}        &\ding{51} (MLM)          &                &              &                 &                       \\
Tapex \cite{liu2021tapex}      &                       &                &\ding{51}    &                 &                       \\
TableFormer \cite{yang2022tableformer} &\ding{51} (MLM)        &                &              &                 &                       \\
% UnifiedSKG  &                     &               &          & \ding{51}      &               \\
SCORE \cite{yu2020score}      &\ding{51} (MLM)           &\ding{51} (CCS)     &              &                 &\ding{51} (TCS)      \\
\bottomrule
\end{tabular}
\label{tab:pretrain-objectives}
\end{table*}

\subsection{Pre-training Objectives}
Most existing pre-training models for text-to-SQL parsing employ either a single Transformer or a Transformer-based encoder-decoder framework as the backbone, and adopt different kinds of pre-training objectives to capture the characteristics of the text-to-SQL parsing task. 
As illustrated in Table \ref{tab:pretrain-objectives}, the pre-training objectives can be divided into  five primary categories, including masked language modelling (MLM) \cite{yin2020tabert,herzig-etal-2020-tapas,yu2020grappa,shi2021learning,eisenschlos2021mate,yang2022tableformer,yu2020score}, schema linking \cite{yu2020grappa,shi2021learning,deng2020structure,yu2020score}, SQL executor \cite{liu2021tapex}, text generation \cite{shi2021learning} and context modelling \cite{yu2020score}. Next, we will introduce the implementation details of each primary pre-training objective. 

%And many recent works, such as: TaBERT, TAPAS, GRAPPA, GAP, STRUG, MATE, Tapex, TableFormer, and SCORE, propose different pre-training objectives to enhance the different characteristics of the text-to-SQL parsing task. For example, TaBERT proposed masked language modeling (MLM), Masked Column Prediction (MCP) and Cell Value Recovery (CVR); TAPAS \cite{herzig-etal-2020-tapas} and TableFormer \cite{yang2022tableformer} proposed MLM with whole word masking on text and whole cell masking on table cell; GRAPPA proposed SQL semantic prediction (SSP); GAP proposed column recovery task (CPred) and incomplete SQL generation task (GenSQL); STRUG proposed column grounding (CG) and value grounding (VG); MATE proposed MLM objective applied to the corpus of tables and text; Tapex proposed a SQL Executor; SCORE proposed turn contextual switch objective (TCS) and Column Contextual Semantics (CCS). The detailed aspects of different pre-training objectives are introduced in the following section.

%\yangmin{I believe the categories of pre-training objectives should be consistent with Figure 1.}

%Most pre-training objectives of table pre-training model generally follow the Masked Language Modeling (MLM), which mask a token and recovery it. While a variety of other pre-training objectives mimic downstream tasks.

\subsubsection{Masked Language Modelling}
Existing works often explore different variants of masked language modeling (MLM) to guide the language models to learn better representations of both natural language and tabular data. Concretely, the MLM objectives can be divided into three primary categories: reconstructing the corrupted NL sentences, reconstructing the corrupted table headers or cell values, and reconstructing the tokens from the corrupted NL sentences and tables. 

In particular, most pre-training models \cite{yin2020tabert,herzig-etal-2020-tapas,yu2020grappa,shi2021learning,eisenschlos2021mate,yang2022tableformer,yu2020score} adopt masked language modelling by randomly masking a part of the input tokens from NL sentences or table headers and then predicting the masked tokens. Then, the MLM loss is calculated by minimizing the cross-entropy loss between the original masked tokens and the predicted masked tokens. 
In addition, \textsc{TaBERT} \cite{yin2020tabert} also proposed a Masked Column Prediction (MCP) and a Cell Value Recovery (CVR) to learn the column representations of tables, where the MCP objective  predicts the names and data types of masked columns and the CVR objective attempts to predict the original value of each cell in the masked column given its cell vector. \textsc{GAP} \cite{shi2021learning} devised a Column Recovery (CRec) objective to recovery the corresponding column name conditioned on a sampled cell value.

\subsubsection{Schema Linking}
Schema linking is a key component in text-to-SQL parsing, which learns the alignment between NL questions and given tables. In particular, schema linking aims at identifying the references of columns, tables and condition values in NL questions. It is especially important for complex SQL generation and cross-domain generalization, where the text-to-SQL parser should be aware of what tables and columns are involved in the NL question even when referring against with tables from arbitrary domains and modelling complex semantic dependencies between NL questions and SQL queries.  

Recently, several pre-training objectives \cite{yu2020grappa,shi2021learning,deng2020structure,yu2020score} are devised to model the schema linking information by learning the correlations of NL questions and tables. \textsc{GraPPa} \cite{yu2020grappa} proposed a SQL Semantic Prediction (SSP) objective, which aims at predicting whether a column name appears in the SQL query and which SQL operation is triggered conditioned on the NL question and given table headers. The SSP objective is implemented by converting the SQL sequence labeling into operation classification for each column, which results in 254 possible operation classes.
Similar to \textsc{GraPPa}\cite{yu2020grappa},  \textsc{SCoRE} \cite{yu2020score} proposed a Column Contextual Semantics (CCS) objective, aiming to predict what operation should be performed on the given column. 
\textsc{STRUG} \cite{deng2020structure} proposed three structure-grounded objectives to learn the text-table alignment, including Column Grounding (CG), Value Grounding (VG) and Column-Value mapping (CV). Concretely, the CG objective is a binary classification task, aiming at predicting whether a column is mentioned in the NL question or not. The VG objective is also transformed to  binary classification, which aims at predicting where a token is a part of a grounded value conditioned on the NL question
and the table schema. To further align the grounded columns and values, the CG objective is devised to match the tokens in the NL question and the columns.
Similar to the CG objective, \textsc{GAP} \cite{shi2021learning} also developed a Column Prediction (CPred) objective to predict whether a column is used in the NL question or not.

\subsubsection{SQL Executor}
Modelling structured tables plays a crucial role in text-to-SQL pre-training. \textsc{TaPEx} \cite{liu2021tapex} proposed a SQL executor objective by pre-training a neural model to mimic a SQL executor on tables.  Specifically, the neural SQL executor is learned to execute the SQL query and output the corresponding correct result, which requires the model to have deep understanding of the SQL queries and tables. 

%\textsc{TaPEx} \cite{liu2021tapex} proposes a pre-training objective which encourages a language model to be a neural SQL executor. Given a table and an executable SQL query, it obtains the query’s execution result through an off-the-shelf SQL executor (e.g., MySQL) to serve as the supervision for the model. They think if a language model can be trained to faithfully “execute” SQL queries and produce correct results, then it should have a deep understanding of tables.

\subsubsection{SQL Generation}
The goal of text-to-SQL parsing is to translate NL questions into SQL queries that can be executed on the given tables. Therefore, it can be beneficial to incorporate the SQL generation objective into the pre-training methods so as to further enhance the downstream tasks. \textsc{GAP} \cite{shi2021learning} proposes a SQL generation objective to generate specific SQL keywords or column names in the appropriate positions rather than merely predict whether a column is mentioned or not.

%As mentioned above, text-to-SQL parsing, the important intermediate task of table QA, aims to parse natural language queries to SQLs (text-to-SQL), logical forms, or other kinds if programming languages that can return answers through executing on one or more target tables. Therefore, some works directly use the SQL generation as the pre-training objective to enhance the downstream tasks.

%GAP proposes an incomplete SQL generation task (GenSQL) which is slight different from downstream task. Specifically, the GAP decoder emits the target SQL token by token with a close vocabulary set, which is composed of the SQL keywords vocabulary and column names. At each decoding step, the decoder generates a hidden vector and then a dot- product operation is applied on it and the target vocabulary representations, yielding a probability distribution over the vocabulary set.

\subsubsection{Turn Contextual Switch}
The above pre-training objectives primarily model stand-alone NL questions without considering the context-dependent interactions, which result in sub-optimal performance for context-dependent text-to-SQL parsing. \textsc{SCoRe} \cite{yu2020score} is the first representative pre-training method for context-dependent Text-to-SQL parsing. In particular, \textsc{SCoRe} designs a turn contextual switch (TCS) objective to model the context flow by predicting the context switch label (from 26 possible operations) between two consecutive user utterances.
Despite its effectiveness, the TCS objective ignores the complex interactions of context utterances, and it is difficult to track the dependence between distant utterances. 

\section{Future Directions}
Despite the remarkable progress of previous methods, there remain several challenges for developing high-quality text-to-SQL parsers.  Based on the works in this manuscript, we discuss several directions for future exploration in the field of text-to-SQL parsing. 

\subsection{Effective High-quality Training Data Generation}
The current benchmark datasets for text-to-SQL parsing are still limited by the quality, quantity and diversity of training data. For instance, WikiSQL \cite{zhong2017seq2sql}  contains a large amount of simple question-SQL pairs and single tables, which neglects the quality and diversity of the training instances. In particular, WikiSQL \cite{zhong2017seq2sql} has several limitations which are described as follows. 
First, the training, development and testing sets in  WikiSQL \cite{zhong2017seq2sql} share the same domains, without concerning the cross-domain generalization ability of the text-to-SQL parsers.  
Second, the SQL queries in WikiSQL \cite{zhong2017seq2sql} are simple, which do not contain complex operations, such as ``ORDER BY'', ``GROUP BY'', ``NESTED'' and ``HAVING''.
Third, each database in WikiSQL \cite{zhong2017seq2sql} contains only one table, simplifying the text-to-SQL parsing. 
Spider \cite{yu2018spider} is a  cross-domain and complex benchmark dataset for text-to-SQL parsing, which contains complicated SQL queries and databases with multiple tables in different domains. Spider \cite{yu2018spider} is proposed to investigate the ability of a text-to-SQL parser to generalize to not only new SQL queries and  database schemas but also new domains. However, Spider \cite{yu2018spider} merely has about ten thousand samples, which is not large enough to build a high-quality text-to-SQL parser. 
There are also many data generation methods that apply rule-based methods to produce a large amount of question-SQL pairs. However, these automatically generated samples are of inferior quality and usually lose diversity.
Therefore, how to construct text-to-SQL corpora with high quality, large-scale quantity and high diversity is an important future exploration direction.

\subsection{Handling Large-scale Table/Database Morphology}
%Many approaches for text-to-SQL parsing are generally light-weight algorithms which can be transmitted across the proposed networks \yangmin{What's the meaning of ``which can be transmitted across the proposed networks''?}. 
% The usage of database schemas has been restricted to table names, column names and a few cell values for small-scale databases. However, in real-world scenarios, there is a huge amount of data stored in real-world databases while current text-to-SQL models are hardly able to solve.
The tables used in current benchmark corpora usually contain less than ten rows and columns by considering the input length limitation of current neural text-to-SQL models. However, in many real-world applications, the involved tables usually consist of thousands of rows and columns, which pose a big challenge to the memory and computational efficiency of existing neural text-to-SQL models. In particular, when the number or size of involved tables becomes too large, how to encode the table schemas and retrieve appropriate knowledge from large tables are challenging. Thus, more future efforts should be made on how to (i) develop effective text-to-models that can encode a long sequence of table schemas, and (ii) improve the execution efficiency of the generated SQL when involving a large-scale database.

\subsection{Structured Tabular Data Encoding}

Different from textual data, the tabular data involved in text-to-SQL parsing is distributed in 2-D structures. Most text-to-SQL parsing methods first convert the 2-D table data into the linearized 1-D sequence, which is then fed into the input encoder. Such linearization methods cannot capture the structural information of tables. In addition, most previous works primarily focus on web tables, while more other kinds of tabular data that contain hyperlinks, visual data, spreadsheet formulas, and quantities are not considered. It is non-trivial to learn high-quality representations from such diverse data types by directly encoding a linearized input sequence. 
It is worth exploring text-to-SQL methods for effectively encoding structural information of such 2-D tabular data, so that more comprehensive input representations can be learned to facilitate the SQL generation. 

%Some text-to-SQL methods such as \textsc{Tapas} \cite{herzig-etal-2020-tapas} and MATE \cite{eisenschlos2021mate} encode the column and row representations based on column/row IDs. TableFormer \cite{yang2022tableformer} attempts to capture the structural information of the tabular data with the column/row relations similar to the relative positions between two linearized tokens.

%The basic idea of encoding table now is to linearize the table content and encode it as normal text. But a key point is table has 2-D structure, which are arranged in two-dimension: row and column. Such linearization method would destroy the structural information of tables. Besides, most works only focus on a specific type of table, e.g., web tables, in fact, more kinds of tables includes not only text but also quantities, visual formats, hyperlinks, and even spreadsheet formulas. Directly encoding a flat of sequences makes it non-trivial to learn high-level representations from such diverse data types. Works such as Tapas \cite{herzig-etal-2020-tapas} and MATE \cite{eisenschlos2021mate} learn column and row encodings based on column/row IDs. TableFormer \cite{yang2022tableformer} captures the structural information with a same column/row relation as one kind of relative position between two linearized tokens. But most works still apply linearization method to input table content, while destroy the structure of table. How to encode structural information of such 2-D structure table and its special data type is an urgent challenge.

\subsection{Heterogeneous Information Modeling}
Existing text-to-SQL datasets, such as Spider \cite{yu2018spider}, mainly contain textual and numeric data from NL questions and tables.
However, many real-life applications contain more types of data (e.g., images) over heterogeneous forms. Using only homogeneous information source cannot satisfy the demands of some real-life applications, such as E-commerce and metaverse. For example, \cite{talmor2021multimodalqa} takes the first step towards this direction by proposing a challenging question answering dataset that requires joint reasoning over texts, tables and images. 
When applying the text-to-SQL models in the E-commerce and metaverse domains, it is necessary to process multimodal data and aggregate information from heterogeneous information sources for obtaining the correct returned results. 

%\qinbowen{%However, some real applications contain distributed knowledge over heterogeneous forms, using only homogeneous information source might not coverage boarder conditions, such as E-commerce and metaverse. In the future, in the e-commerce of meta-verse, when querying items, it is possible to combine knowledge of multiple modalities to search the database. Therefore, the heterogeneous information based text-to-SQL is an urgent need for real application. Basically, natural language can be categorized as free-form text and structured text. The former refers to large text corpus, while the latter refers to different kinds of tables, knowledge bases and knowledge graphs. \cite{chen2020hybridqa} points that the free-form corpus coverage different knowledge better while structured data has better compositionality to handle complex multi-hop questions. Therefore, to better simulating realistic setting, different knowledge representation forms should be combined and the model requires to aggregate information from heterogeneous information sources for answering a question. Even more, in some multi-modal settings, knowledge may be distributed into text, image, video or speech. A realistic question should be answered by multiple pieces of evidence from visual, textual and tabular sources.  \cite{talmor2021multimodalqa} took the first step towards this direction by proposing a challenging question answering dataset that requires joint reasoning over text, tables and images. It is an urgent problem how to define heterogeneous information based text-to-SQL parsing.}

\subsection{Cross-domain Text-to-SQL Parsing}
Most existing works have worked on in-domain text-to-SQL parsing, where the training and testing sets share the same domains. 
However, no matter how much data is collected and applied to train a text-to-SQL parser, it is difficult to cover all possible domains of databases. Thus, when deployed in practice, a well-trained text-to-SQL parser cannot generalize to new domains and often performs unsatisfactorily.
Although some cross-domain text-to-SQL datasets (e.g., Spider \cite{yu2018spider}, DK \cite{spiderDK} and SYN \cite{spiderSYN}) have been constructed for the challenging cross-domain settings, there are no text-to-SQL methods that design tailored algorithms to deal with the out-of-distribution (OOD) generalization problem where the testing data distribution is unknown and different from the training data distribution.
The supervised learning method is fragile when being exposed to data with different distributions. %The ability to generalize under the OOD generalization is of essential significance. 
Therefore, how to explore the OOD generalization of the text-to-SQL parser is a promising future direction for both academic and industry communities. 
%While there has been much progress in text-to-SQL, earlier work has primarily focused on evaluating parsers in-domain problem, while rare considers more realistic situation: achieving domain generalization. Spider \cite{yu2018spider} benchmark firstly proposed a challenging settings: cross-domain, which requires a system to generalize to new domains. The train, dev and test set of spider benchmark do not share the database and tables, which can be seen as a zero-shot domain problem. However, few or no learning algorithms or goals focus on promoting domain generalization, and almost all existing methods rely on standard supervised learning. Traditional supervised learning only assumes that the source and target domain data come from the same distribution, so it is difficult to capture the concept of domain generalization for zero-shot semantic parsing. \cite{wang2020meta} firstly proposed a meta-learning framework which targets zero-shot domain generalization for semantic parsing. The experimental results shows that it can generalize to unseen domain compared to the supervised learning goal. From the perspective, the future text-to-SQL system should explore more challenging cross-domain settings.

\subsection{Robustness of Text-to-SQL Parsing Models} 
The robustness of text-to-SQL parsing models pose a prominent challenge when being deployed in real-life applications. Small perturbations in the input may significantly reduce the performance of text-to-SQL parsing models.  High performance requires robust performance on noisy inputs. Recently, Spider-SYN \cite{spiderSYN} investigates the robustness of text-to-SQL parsing models to synonym substitution by removing the explicit schema-linking correspondence between NL questions and table schemas.
Spider-DK \cite{spiderDK} investigates the generalization of text-to-SQL parsing models by injecting rarely observed domain knowledge into the NL questions so as to evaluate the model understanding of domain knowledge. 
The experimental results from both \cite{spiderSYN} and \cite{spiderDK} demonstrated that the performance of text-to-SQL models are inferior when facing the  small perturbations (synonym substitution and domain knowledge injection) in the input, even though the training and test data shares similar distributions. 
Hence, it is necessary to stabilize the neural text-to-SQL parsing models, making the models more robust to different perturbations
There is very few exploration for improving the robustness of text-to-SQL models, thus more efforts should be paid to this research direction.

%Recently, there has been significant progress in studying neural networks for translating text descriptions into SQL queries under the zero-shot cross-domain setting. Despite achieving good performance on some public benchmarks, such as spider benchmark \cite{yu2018spider}, some works observe that existing text-to-SQL models do not generalize well on some special cases. For example, \cite{spiderSYN} found that accuracy dramatically drops by eliminating the explicit schema-linking correspondence between NL questions and table schemas. \cite{spiderDK} observed that existing text-to-SQL models do not generalize when facing do-main knowledge that does not frequently appear in the training data, which may render the worse prediction performance for unseen domains. Therefore, it is important to investigate the robustness of current text-to-SQL models. \cite{spiderSYN} proposed two approach to enhance the robustness of text-to-SQL models: the first category of approaches utilizes additional synonym annotations for table schemas by modifying the model input, while the second category is based on adversarial training. \cite{hui2022s} injecting syntax to question-Schema graph encoder for text-to-SQL parsers, which effectively leverages the syntactic dependency information of questions in text-to-SQL to improve the robustness of current text-to-SQL models. 

\subsection{Zero-shot and Few-shot Text-to-SQL Parsing}
The ``pre-training+fine-tuning'' paradigm has been widely used for text-to-SQL parsing, yielding state-of-the-art performances. Although the text-to-SQL parsing models that are trained via the ``pre-training+fine-tuning'' paradigm can obtain impressive performance, they still require a large-scale annotated dataset for fine-tuning the downstream tasks.
These neural models could show poor performance in zero-shot setting without any task-specific annotated training data.
One possible solution is to adopt pre-trained language models (e.g., GPT-3 \cite{floridi2020gpt} and Codex \cite{chen2021evaluating}) for zero-shot transfer to downstream tasks without the fine-tuning phase. 
\cite{rajkumar2022evaluating} and \cite{xie2022unifiedskg} revealed that large pre-trained language models can achieve competitive performance on text-to-SQL parsing without fine-tuning. For example, Codex \cite{chen2021evaluating} achieved the execution accuracy up to 67\% on the Spider \cite{yu2018spider} development set. Therefore, zero-shot and few-shot text-to-SQL parsing is a promising direction for future exploration. 

%An approach to text-to-SQL task involves training a model to produce a SQL query when given a question, a database schema, and possibly database content as inputs. With the development of large language models, a clear trend in this area is to fine-tune large language models. Recent results from the broader field demonstrate that simply scaling training data and model size for generative language models brings advanced capabilities, such as few-shot learning without finetuning (GPT-3 and Codex). Therefore, it is interesting to investigate if such large language models are already competitive Text-to-SQL solutions without any further finetuning on task-specific training data.
%\cite{rajkumar2022evaluating} and \cite{xie2022unifiedskg} proved that large language model can have a competitive performance without finetuning, such as Codex achieves a performance of up to 67\% execution accuracy on the Spider development set.

\subsection{Pre-Training for Context-dependent Text-to-SQL Parsing}
Context-dependent text-to-SQL parsing need to effectively process context information so as to effectively generate the current SQL query since users may omit previously mentioned entities as well as constraints and introduce substitutions to what has already been stated. The key challenge is how to track and explore the interaction states in history utterances to assist the models to better understand the current NL utterance. Nevertheless, most prior TaLMs primarily model stand-alone NL utterances without considering the context-dependent interactions, which result in sub-optimal performance. 
Although \textsc{SCoRe} \cite{yu2020score} 
model the turn contextual switch by predicting the context switch label between two consecutive user utterances, it ignores the complex interactions of context utterances and cannot track the dependence between distant utterances.
Inspired by the dialogue state tracking \cite{wang2021tracking,he2022unified,he2022galaxy} which keeps track of user intentions in the form of a set of dialogue states (e.g., slot-value pairs) in task-oriented dialogue systems, it is  beneficial to keep track of schema states (or user requests) of context-dependent SQL queries in the pre-training process. How to explore and make good use of context information to enrich question and schema representations is critical for the deployment of text-to-SQL generalization methods.

%Task-oriented dialogue systems rely on pre-defined slots and values for request processing, such as flight or hotel booking or transportation planning. Those pre-defined slots and values are called dialogue states. In fact that dialogue states are used to determine how to query backend databases for the information users need. Cosql \cite{yu2019cosql} pointed out that rather than grounding the dialogue states by domain-specific slot value pairs, the dialogue states can also be grounded in SQL, a domain-independent executable representation. Representing dialogue states by slot-value pairs results in that these systems only operate on a small number of domains and have difficulty capturing the diverse semantics of practical user questions. Therefore, represented the dialogue states using simple SQL queries consisting of SELECT and WHERE clauses) can support general-purpose exploration and querying of databases by end users. Cosql \cite{yu2019cosql} and IC-DST \cite{ic-dst} both convert the dialogue states to SQL queries, where the former proposed a SQL-ground dialogue states tasks, and the latter reformulated DST into a text-to-SQL problem. For example, when the pre-defined slot-value pairs comes to "restaurant-food: French", it can be changed to SQL format: "SELECT * FROM restaurant WHERE food=French".

\subsection{Interpretability of Text-to-SQL Models}
Deep neural networks (DNNs) have achieved the state-of-the-art performance on text-to-SQL parsing. However, the neural models usually lack of interpretability and are perceived as black boxes. For the sensitive domains such as finance or healthcare, the interpretability of neural text-to-SQL models and returned results is necessary, which is as important as the SQL generation performance. Instead of exploring the implicit encoding and reasoning strategies making the predicted results uninterpretable to humans, how to take advantage of both the representation ability of DNNs and the explicit reasoning ability of symbolic approaches is a promising future direction for text-to-SQL parsing, especially for the applications in the highly sensitive domains.

%Another important problem is reasoning ability when requiring knowledge outside the database to answer. In \cite{yu2018spider}, the authors mentioned humans sometimes ask questions that require common sense knowledge outside the given database. For example, when question ask “X reports to Y”, it actually corresponds to an “employee-manager” relation. In the future, it is highly desirable to focus on the ability of common sense inference and math calculation.

\subsection{Data Privacy-preserving}
Many text-to-SQL models have been deployed on the cloud to process user data by the small businesses.
Due to the sensitivity of user data, data privacy-preserving could be an essential but challenging task in text-to-SQL parsing. In particular, most text-to-SQL parsing models require identifying patterns as well as relations from large-scale corpora and related database schemas and gathering all information into a central site for feature representation learning. However, the data privacy issue may prevent the neural models from building a centralized architecture given the fact that the relations or patterns among different database schemas may be distributed among several custodians, none of which are allowed to transfer their user data to other sites. 
How to effectively transform the NL questions as well as database schema to privatized input representations on the local devices and then upload the transformed representations to the service provider would be also a promising exploration direction. 

%As mentioned in the previous sections, many approaches applied in text-to-SQL parsing require identifying patterns and relations from large quantities of data and related schemas and gathering all information into a central site for feature modeling. However, in real-world scenarios, privacy concerns can prevent building a centralized architecture given the fact that relations or patterns among different schemas may be distributed among several custodians, none of which are allowed to transfer their data to another site. There are many variants of this problem, depending on how the data is distributed, what type of information mining we wish to do, and what restrictions are placed on sharing of information. Some problems are quite tractable, others are more difficult. Therefore, we assume that future researches can focus on addressing the problem of computing association rules within such scenario for text-to-SQL parsing over databases under certain restrictions.
\section{Conclusion}
This manuscript aims at providing the first comprehensive survey for text-to-SQL parsing from the NLP community. We provide an overview of the existing datasets, current neural text-to-SQL parsing models,  tailored pre-training methods, and possible future exploration directions. First, we introduce available single-turn and multi-turn text-to-SQL corpora and provide a tabular to summarize these resources. Second, we review the state-of-the-art neural text-to-SQL parsing models based on varying encoders and decoders.  Third, we present the well-known tabular pre-training methods for text-to-SQL parsing and elaborate on existing pre-training objectives. Finally, we discuss the future directions in the field of text-to-SQL parsing. We hope that this survey can provide some insightful perspectives and inspire the widespread implementation of real-life text-to-SQL parsing systems.

% use section* for acknowledgment
\ifCLASSOPTIONcompsoc
  % The Computer Society usually uses the plural form
  \section*{Acknowledgments}
\else
  % regular IEEE prefers the singular form
  \section*{Acknowledgment}
\fi

Min Yang was partially supported by National Natural Science Foundation of China (No. 61906185), Shenzhen Basic Research Foundation (No. JCYJ20210324115614039 and No. JCYJ20200109113441941), Shenzhen Science and Technology Innovation Program (Grant No. KQTD20190929172835662), Youth Innovation Promotion Association of CAS China (No. 2020357). This work was supported by Alibaba Group through Alibaba Innovative Research Program.

\ifCLASSOPTIONcaptionsoff
  \newpage
\fi

% trigger a \newpage just before the given reference
% number - used to balance the columns on the last page
% adjust value as needed - may need to be readjusted if
% the document is modified later
%\IEEEtriggeratref{8}
% The "triggered" command can be changed if desired:
%\IEEEtriggercmd{\enlargethispage{-5in}}

% references section

% can use a bibliography generated by BibTeX as a .bbl file
% BibTeX documentation can be easily obtained at:
% http://mirror.ctan.org/biblio/bibtex/contrib/doc/
% The IEEEtran BibTeX style support page is at:
% http://www.michaelshell.org/tex/ieeetran/bibtex/
\bibliographystyle{IEEEtran}
% \bibliographystyle{IEEEtranN}
% argument is your BibTeX string definitions and bibliography database(s)
\bibliography{IEEEfull}
\end{document}